\documentclass[english]{article}
\usepackage{geometry}
\usepackage[T1]{fontenc}
\usepackage{microtype}
\usepackage{graphicx}
\usepackage{booktabs}
\usepackage{amsmath}
\usepackage{amssymb}
\usepackage{algorithm}
\usepackage{algorithmic}
\usepackage{mathtools}
\usepackage{amsthm}
\usepackage{bbding}
\usepackage{multirow}
\usepackage{xspace}
\usepackage{hyperref}
\usepackage[capitalize,noabbrev]{cleveref}
\usepackage{natbib}
\renewcommand{\cite}{\citep}
\def\eqref#1{equation~\ref{#1}}

\usepackage{amsmath,amssymb,amsthm,bm}

\def\eqref#1{equation~\ref{#1}}
\def\1{\bm{1}}

\DeclareMathAlphabet{\mathsfit}{\encodingdefault}{\sfdefault}{m}{sl}
\SetMathAlphabet{\mathsfit}{bold}{\encodingdefault}{\sfdefault}{bx}{n}

\theoremstyle{plain}
\newtheorem{theorem}{Theorem}[section]

\newtheorem{lemma}[theorem]{Lemma}

\theoremstyle{definition}

\newtheorem{assumption}[theorem]{Assumption}
\theoremstyle{remark}

\usepackage{hyperref}
\usepackage{url}

\usepackage[T1]{fontenc}
\usepackage{algorithmic}
\usepackage{algorithm}
\usepackage{tcolorbox}
\usepackage{graphicx}
\usepackage{bbding}
\usepackage{subcaption}
\usepackage{booktabs}
\usepackage{multirow}
\usepackage{titletoc}
\renewcommand{\cite}{\citep}
\usepackage{color, colortbl}
\definecolor{blue1}{rgb}{1, 1, 1}
\definecolor{blue2}{rgb}{1, 1, 1}
\definecolor{blue3}{rgb}{1, 1, 1}
\definecolor{blue4}{rgb}{1, 1, 1}

\definecolor{DarkPink}{rgb}{0.5,0.0,0.18}
\definecolor{DarkGreen}{rgb}{0.1,0.5,0.1}
\definecolor{DarkRed}{rgb}{0.5,0.1,0.1}
\definecolor{DarkBlue}{rgb}{0.1,0.1,0.7}
\definecolor{DarkYellow}{rgb}{.79,.79,0}
\hypersetup{
    unicode=false,          % non-Latin characters in AcrobatÂ¿s bookmarks
    pdftoolbar=true,        % show Acrobat toolbar?
    pdfmenubar=true,        % show Acrobat menu?
    pdffitwindow=false,      % page fit to window when opened
    pdfnewwindow=true,      % links in new window
    colorlinks=true,       % false: boxed links; true: colored links
    linkcolor=DarkBlue,          % color of internal links
    citecolor=DarkRed,        % color of links to bibliography
    filecolor=DarkRed,      % color of file links
    urlcolor=DarkBlue,          % color of external links
    pdftitle={},
    pdfauthor={},
}

\begin{document}

\title{Wanda-Guided Sparse Perturbations for \\
Efficient Zeroth-Order Optimization}
\author{
  Ziqing Wen$^{1,*}$,
  Xiaohui Song$^{2}$,
  Jiahuan Wang$^{1}$,
  Ping Luo$^{1}$,
  Junlin Zeng$^{1}$, \\
  Dongsheng Li$^{1}$,
  Tao Sun$^{1,\dag}$
}
\date{}

\maketitle

% Internship contribution
\renewcommand*{\thefootnote}{\fnsymbol{footnote}}
\footnotetext[1]{Work done during an internship at OPPO AI-Center.}

% Affiliations
\renewcommand*{\thefootnote}{\arabic{footnote}}

\footnotetext[1]{National University of Defense Technology. Contact: \texttt{suntao.saltfish@outlook.com}}
\footnotetext[2]{OPPO AI center}

\begin{abstract}
Zeroth-order (ZO) optimization enables memory-efficient fine-tuning of large language models through forward-only evaluations, but high-dimensional random perturbations can limit optimization efficiency. Sparse perturbations reduce the search dimension, making the selection of informative coordinates critical. We propose Wanda-Importance Sparsity-MeZO (WIS-MeZO), which uses Wanda weight--activation saliency to construct task-conditioned sparse perturbation masks while keeping the pretrained model dense. We analyze support quality through gradient energy capture and relate it to gradient estimation and convergence. Across 16 benchmarks with LLaMA and Qwen models ranging from 3B to 32B parameters, WIS-MeZO achieves higher aggregate scores than MeZO and reaches comparable validation losses in fewer optimization steps. Comparisons with random and magnitude-based sparse perturbations further highlight the importance of informative support selection.
\end{abstract}

\section{Introduction}

Fine-tuning large language models (LLMs) requires substantial memory for the activations, gradients, and optimizer states used in backpropagation. Parameter-efficient fine-tuning~\cite{hu2022lora,liu2024dora}, quantization~\cite{dettmers2022gpt3_int8}, and memory-efficient optimizers~\cite{zhao2024galore,zhu2025apollo} reduce different parts of this cost. Zeroth-order (ZO) optimization offers another approach by estimating update directions through forward evaluations, as demonstrated by MeZO~\cite{malladi2023_mezo}. Its reduced training-memory requirements are particularly appealing for on-device adaptation, where models must be personalized under limited device resources~\cite{guo2025zerothorder_szoq}.

Despite this memory advantage, ZO optimization remains challenging in the high-dimensional parameter spaces of LLMs. The variance of random-direction gradient estimates grows with the number of perturbed coordinates, making it difficult to obtain informative updates. Sparse ZO methods address this issue by restricting perturbations to a subset of parameters. However, reducing the number of coordinates is only part of the problem: the selected support must also retain useful information for downstream optimization. Existing methods select small-magnitude weights~\cite{liu2025sparse} or transfer sensitivity information from pre-training~\cite{guo2025zerothorder_szoq}. This motivates selecting sparse supports using information available from the downstream task itself.

We propose Wanda-Importance Sparsity-MeZO (WIS-MeZO), which uses Wanda weight--activation saliency~\cite{wanda} to select task-conditioned perturbation coordinates. A lightweight forward calibration produces a row-wise sparse mask that is reused during MeZO updates, while the pretrained model remains dense. In this way, downstream activation information guides the allocation of a fixed perturbation budget.

To understand the effect of coordinate selection, we study \emph{gradient energy capture}, which measures the proportion of gradient energy contained in the selected support. This perspective connects support quality to the alignment of ZO estimates and to convergence. We complement the analysis with comparisons against alternative sparse supports and evaluations across downstream tasks and model scales.

Our main contributions are as follows.
\begin{itemize}
\item \textbf{Task-Conditioned Sparse Perturbations.} We introduce WIS-MeZO, which selects sparse perturbation coordinates with downstream Wanda saliency while retaining the dense pretrained model and requiring only forward calibration.

\item \textbf{Gradient-Energy Analysis.} We analyze the relationship between saliency-based support selection and captured gradient energy, and establish theoretical results connecting support quality to ZO estimation and convergence.

\item \textbf{Empirical Evaluation.} Experiments across 16 benchmarks and LLaMA and Qwen models up to 32B parameters show improved aggregate performance over MeZO. Comparisons with random and magnitude-based sparse perturbations, together with validation-loss curves, further demonstrate the benefit of informed support selection.
\end{itemize}

\section{Related Work}

Zeroth-order optimization estimates update directions from function evaluations instead of backpropagation~\cite{spall2002multivariate,ilyas2018black}. Classical approaches include ZO-SGD~\cite{zo_sgd} and ZO-Adam~\cite{chen2019zo_adamm}, while MeZO~\cite{malladi2023_mezo} enables memory-efficient LLM fine-tuning through seed-replayed perturbations. Subsequent work improves ZO estimation through variance reduction~\cite{gautam2024variance}, preconditioning~\cite{zhao2025pazo}, additional queries~\cite{dang2026fzoo}, and low-rank or structured perturbations~\cite{zhang2025mazo,yu2025subzero,chen2025lozo}. AGZO~\cite{lin2026agzo} also uses downstream activations, constructing low-rank perturbation subspaces from current mini-batches. A complementary approach is to reduce the search dimension by selecting a sparse set of weight coordinates.

Sparse coordinate selection has been studied in ZO optimization~\cite{ohta2020sparse,gu2021efficient,cai2022zeroth,zhang2024revisitzo}. DeepZero~\cite{chen2023deepzero} exploits finite-difference information, Sparse-MeZO~\cite{liu2025sparse} selects small-magnitude weights, and SensZOQ~\cite{guo2025zerothorder_szoq} uses sensitivity information from pre-training. Separately, Wanda~\cite{wanda} and SparseGPT~\cite{frantar2023sparsegpt} study weight importance for pruning, while dynamic sparse training~\cite{evci2020rigging,joo2026surprising} adapts sparse networks during optimization. Building on Wanda's weight--activation score, WIS-MeZO uses downstream data to select perturbation coordinates while keeping the model dense. We analyze the quality of this support through gradient energy capture.

\section{Preliminaries}

We first review the MeZO estimator and the Wanda score, which together form the basis of WIS-MeZO.

\subsection{Zeroth-Order Optimization}

For a linear layer, let $\mathbf{W}\in\mathbb{R}^{m\times n}$ denote the weight matrix and let $\mathbf{G}=\nabla f(\mathbf{W},\mathcal{B})$ be its gradient on a training mini-batch $\mathcal{B}$. MeZO~\cite{malladi2023_mezo} estimates this gradient using a random direction $\mathbf{Z}$ and a perturbation radius $\epsilon>0$,
\begin{equation}
\hat{\mathbf{G}}
=\frac{f(\mathbf{W}+\epsilon\mathbf{Z},\mathcal{B})-f(\mathbf{W}-\epsilon\mathbf{Z},\mathcal{B})}{2\epsilon}\mathbf{Z}.
\nonumber
\end{equation}
The entries of $\mathbf{Z}$ are independent standard Gaussian variables. Each estimate uses one direction and two loss evaluations on the same mini-batch. As $\epsilon\to0$, its expectation over $\mathbf{Z}$ approaches $\mathbf{G}$. The update is $\mathbf{W}^{t+1}=\mathbf{W}^{t}-\eta_t\hat{\mathbf{G}}^t$.

The matrix notation describes a single layer with $mn$ coordinates. For full-model updates, $\mathbf{W}$ and the corresponding masks denote the collections of parameter blocks being optimized. Their inner products and squared Frobenius norms sum over these blocks. We use
\begin{equation}
d=\sum_{\ell}m_{\ell}n_{\ell},
\qquad
d_{\mathrm{eff}}=\sum_{\ell}m_{\ell}k_{\ell}
\nonumber
\end{equation}
for the total candidate dimension and the number of selected coordinates, respectively. A common row-wise density $r=k_{\ell}/n_{\ell}$ gives $d_{\mathrm{eff}}=rd$.

\subsection{Wanda Importance Score}

Wanda~\cite{wanda} scores each weight by its magnitude and the activation norm of its input feature. For a weight matrix $\mathbf{W}$ and calibration activations $\mathbf{X}$, the score is
\begin{equation}
\mathbf{S}_{i,j}=|\mathbf{W}_{i,j}|\,\|\mathbf{X}_{j,:}\|_2.
\nonumber
\end{equation}
WIS-MeZO uses this score to select the perturbation mask.

\section{Methodology}

Building on these two ingredients, WIS-MeZO uses Wanda scores to choose perturbation coordinates before applying MeZO updates on the selected support. We describe the support construction and optimization procedure below.

\subsection{WIS-Guided Sparse Perturbation}

For each layer $\mathbf{W}\in\mathbb{R}^{m\times n}$, let $\mathbf{M}\in\{0,1\}^{m\times n}$ be a binary mask. Given a density $r=k/n$, WIS-MeZO selects the $k$ largest scores in each row of $\mathbf{S}$. The mask records which coordinates receive perturbations and updates; the underlying weight matrix remains dense. Row-wise selection allocates the same perturbation budget to every output unit.

The perturbation direction is $\mathbf{U}=(\mathbf{Z}\odot\mathbf{M})/\sqrt{r}$, with independent standard Gaussian entries in $\mathbf{Z}$. For one layer, this gives $\mathbb{E}\|\mathbf{U}\|_F^2=mn$ while restricting the support to $mk$ coordinates. Across the model, the corresponding support dimension is $d_{\mathrm{eff}}$.

\begin{figure*}[!th]
    \centering
    \begin{subfigure}[b]{0.32\linewidth}
        \centering
        \includegraphics[width=\linewidth]{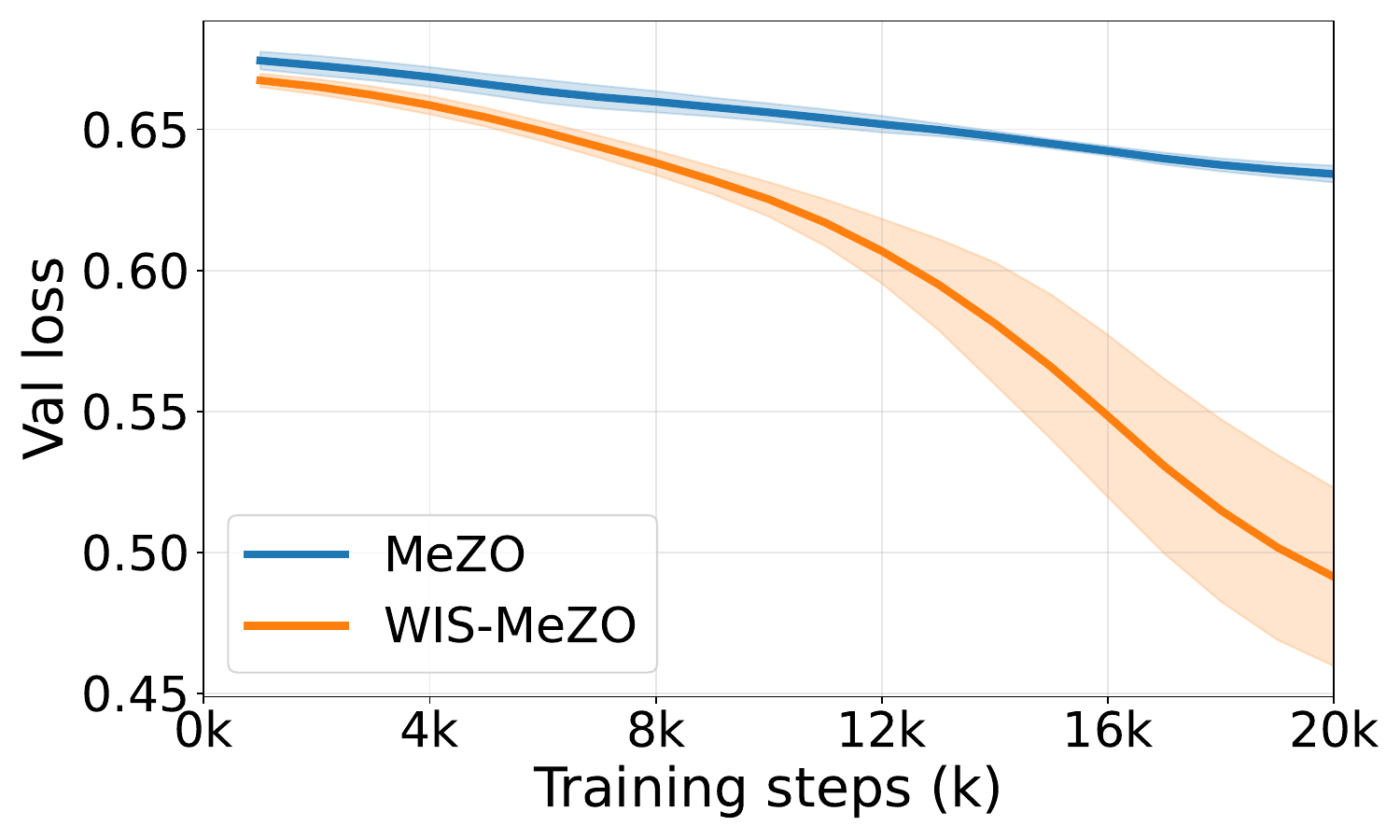}
    \caption{RTE}
    \end{subfigure}
    \hfill
    \begin{subfigure}[b]{0.32\linewidth}
        \centering
        \includegraphics[width=\linewidth]{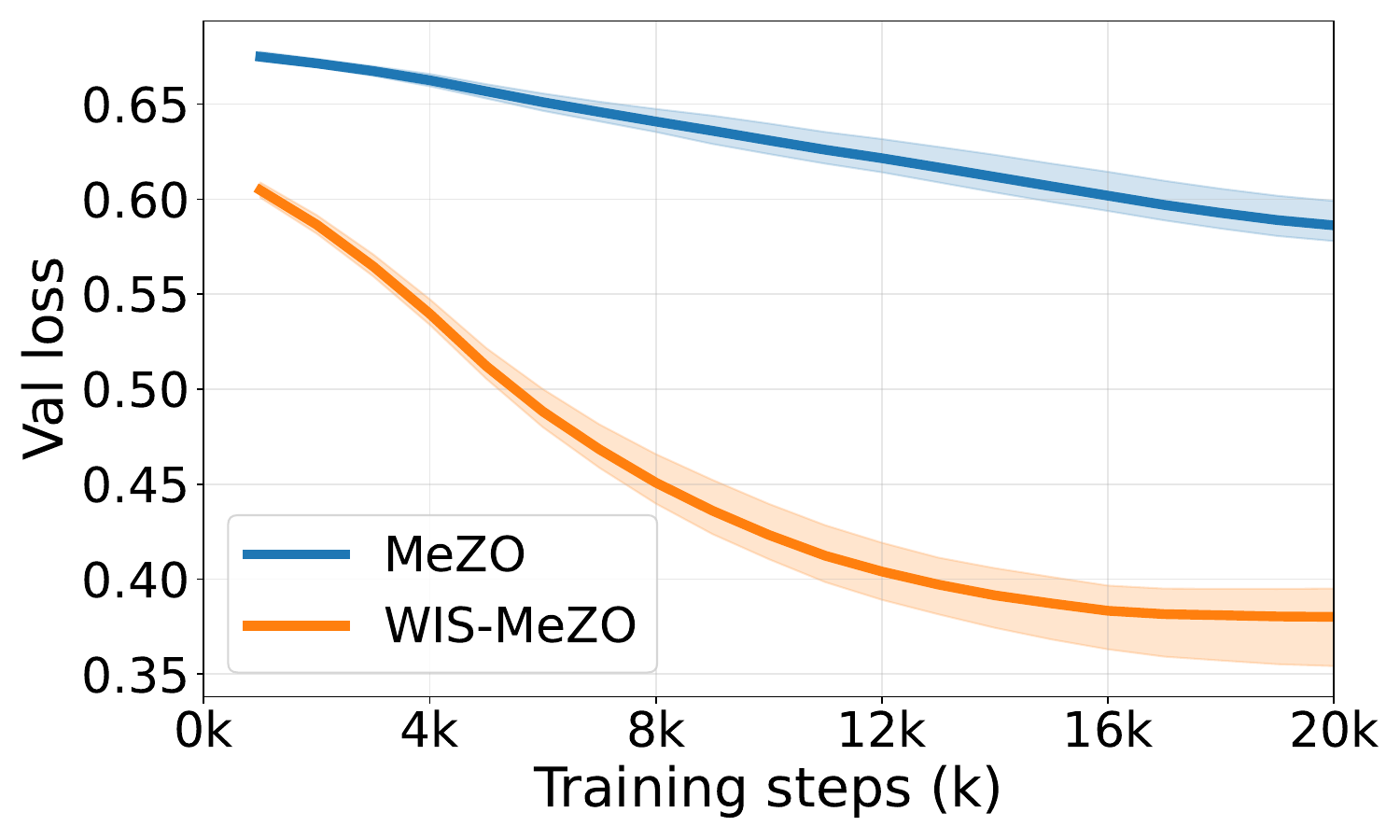}
    \caption{BoolQ}
    \end{subfigure}
    \hfill
    \begin{subfigure}[b]{0.32\linewidth}
        \centering
        \includegraphics[width=\linewidth]{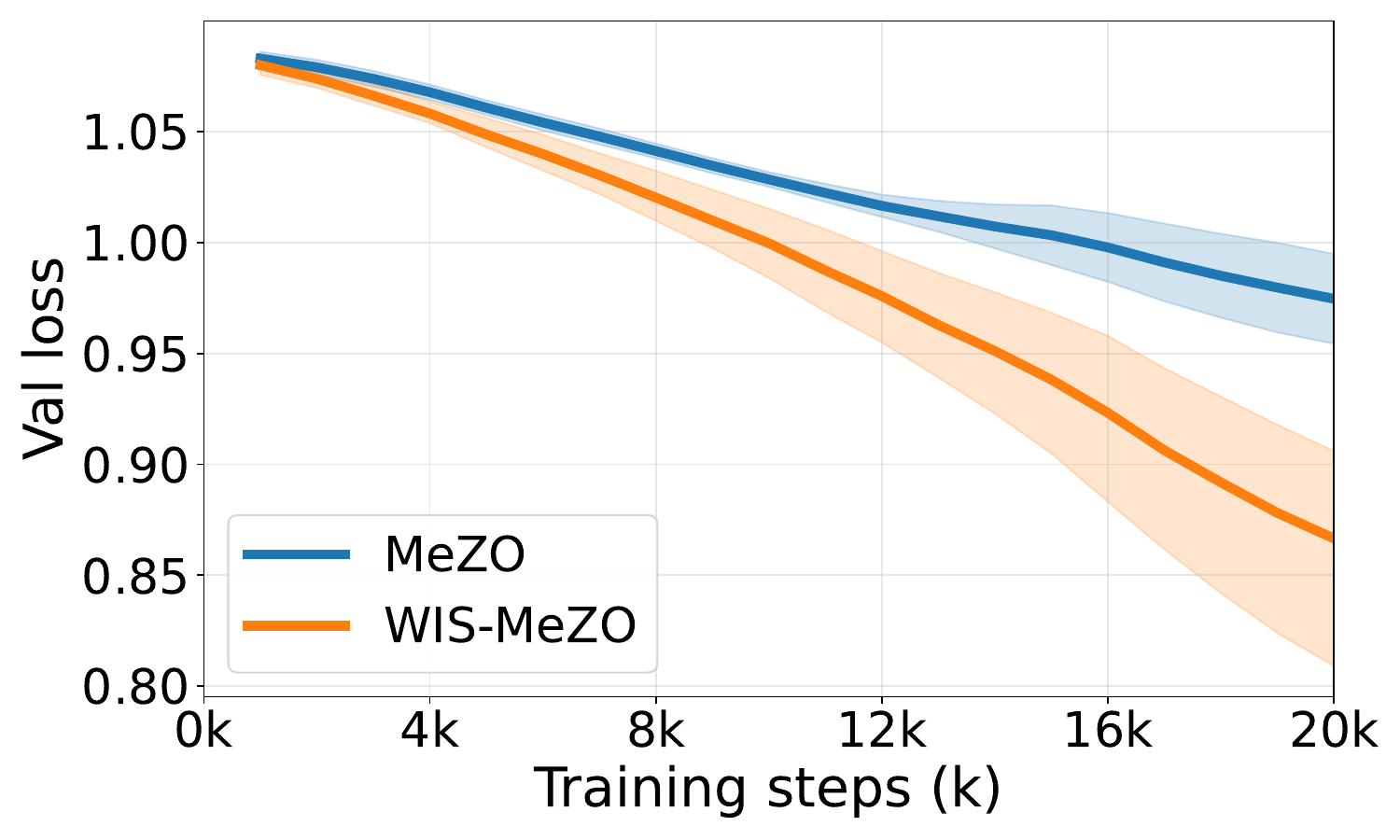}
    \caption{MNLI}
    \end{subfigure}
    
    \caption{\textbf{Validation loss versus training steps for fine-tuning LLaMA3.2-3B, averaged over five random seeds.} WIS-MeZO reaches comparable validation-loss levels in fewer optimization steps.}\label{fig:learning_curve_optimizer}
\end{figure*}

\subsection{Calibration from Downstream Data}
\label{sec:calibration}

WIS-MeZO constructs the mask from 128 examples of the downstream training data using Wanda scores. This calibration is performed once before optimization; WISD-MeZO periodically refreshes the mask every $\tau$ steps. Appendix~\ref{sec:calibration_size} examines the effect and cost of calibration size.

\subsection{The WIS-MeZO Algorithm}
\label{sec:wis-mezo}

In each iteration $t$, we sample a single random seed to generate the
perturbation $\mathbf{Z}^{t}$. The gradient estimator uses one sampled direction
and two loss evaluations on the same training mini-batch,
{\small
\begin{equation}\nonumber
\begin{aligned}
\hat{\mathbf{G}}^{t}
=
\frac{
f\left(
\mathbf{W}^{t}
+
\epsilon\frac{\mathbf{Z}^{t}\odot\mathbf{M}}{\sqrt{r}},
\mathcal{B}
\right)
-
f\left(
\mathbf{W}^{t}
-
\epsilon\frac{\mathbf{Z}^{t}\odot\mathbf{M}}{\sqrt{r}}
,
\mathcal{B}
\right)
}{
2\epsilon
}
\cdot 
\frac{\mathbf{Z}^{t}\odot\mathbf{M}}{\sqrt{r}}.
\end{aligned}
\end{equation}
}
Here, $\odot$ denotes element-wise multiplication, and the update is $\mathbf{W}^{t+1}=\mathbf{W}^{t}-\eta_t\hat{\mathbf{G}}^t$. Seed replay regenerates perturbations for in-place updates without storing full-model gradients. Algorithm~\ref{alg:wis-mezo} summarizes WIS-MeZO, with supplementary pseudocode in Appendix~\ref{app:pseudocode}.

\begin{algorithm}[t]
\caption{WIS-MeZO}
\label{alg:wis-mezo}
\begin{algorithmic}[1]
\STATE \textbf{Input:} Initial weights $\mathbf{W}^{0}$, step sizes $\{\eta_t\}$,
perturbation scale $\epsilon$, mask update frequency $\tau$, density ratio $r$,
calibration dataset $\mathcal{D}_{\mathrm{cal}}$
\FOR{step $t=0,1,2,\dots,T-1$}
    \IF{$t \bmod \tau = 0$}
        \STATE Fetch activations
        $\mathbf{X}\sim\mathcal{D}_{\mathrm{cal}}$
        \hfill \COMMENT{Collect calibration activations}
        \STATE
        $\mathbf{M}\gets
        \operatorname{ComputeMask}(\mathbf{W}^{t},\mathbf{X},r)$
        \hfill \COMMENT{Update mask using Wanda scores}
    \ENDIF
    \STATE Sample training batch $\mathcal{B}$ and random seed $s_{\mathrm{rng}}$
    \STATE
    $\mathbf{W}^{t}\gets
    \operatorname{PerturbParameters}
    (\mathbf{W}^{t},r,\epsilon,\mathbf{M},s_{\mathrm{rng}})$
    \STATE
    $L_{+}\gets f(\mathbf{W}^{t},\mathcal{B})$
    \STATE
    $\mathbf{W}^{t}\gets
    \operatorname{PerturbParameters}
    (\mathbf{W}^{t},r,-2\epsilon,\mathbf{M},s_{\mathrm{rng}})$
    \hfill \COMMENT{Move to the negative perturbation}
    \STATE
    $L_{-}\gets f(\mathbf{W}^{t},\mathcal{B})$
    \STATE
    $\mathbf{W}^{t}\gets
    \operatorname{PerturbParameters}
    (\mathbf{W}^{t},r,\epsilon,\mathbf{M},s_{\mathrm{rng}})$
    \hfill \COMMENT{Recover parameters}
    \STATE Reset the random number generator with seed $s_{\mathrm{rng}}$ and regenerate
    $\mathbf{Z}$
    \STATE
    $\hat{\mathbf{G}}^{t}
    \gets
    \frac{L_{+}-L_{-}}{2\epsilon\sqrt{r}}
    \mathbf{Z}\odot\mathbf{M}$
    \hfill \COMMENT{ZO gradient estimate}
    \STATE
    $\mathbf{W}^{t+1}
    \gets
    \mathbf{W}^{t}-\eta_t\hat{\mathbf{G}}^{t}$
    \hfill \COMMENT{Update the parameters}
\ENDFOR
\end{algorithmic}
\end{algorithm}

With in-place perturbations and seed replay, the main additional storage is the binary mask. A bit-packed representation reduces its cost from 1 byte to 1 bit per candidate parameter, requiring less than 4GB for the 32B model. Together with one-time calibration, this compact storage makes WIS-MeZO a promising approach for memory-constrained on-device adaptation. Section~\ref{sec:memory_speed} reports peak GPU memory and throughput.

\section{Theoretical Analysis}
\label{sec:theoretical_analysis}

Sparse perturbations reduce the search dimension, but different supports retain different amounts of gradient information. We analyze this trade-off through gradient energy capture, first relating Wanda scores to gradient energy and then studying estimator alignment and convergence.

\subsection{Energy Analysis}

For the analysis, let $\mathbf{G}_{i,j}=\delta_i x_j$ be the per-example gradient, with $\delta_i$ denoting the backpropagated signal and $x_j$ an input activation. Write $v_j=\mathbb{E}[x_j^2]$ and $\hat v_j$ for the population and empirical activation second moments.

\begin{assumption}[Bounded Input--Error Dependence]\label{assump:weak_data_corre}
There exist constants $0<\gamma\leq\nu<\infty$ such that
\begin{equation}
\gamma\mathbb{E}[\delta_i^2]v_j
\leq\mathbb{E}[\mathbf{G}_{i,j}^2]
\leq\nu\mathbb{E}[\delta_i^2]v_j.
\nonumber
\end{equation}
\end{assumption}
The constants $\gamma$ and $\nu$ describe the dependence between squared inputs and squared error signals. Independence gives $\gamma=\nu=1$.

\begin{assumption}[Calibration Concentration]\label{assump:sub_gaussian_act}
At the parameter value used for calibration, the $B$ activation observations are independent draws from the downstream distribution. For every input feature,
\begin{equation}
\|x_j\|_{\psi_2}^2\leq K v_j,
\qquad K\geq1,
\nonumber
\end{equation}
where $\|\cdot\|_{\psi_2}$ is the sub-Gaussian norm.
\end{assumption}
Under this assumption, $B=\Omega(K^2\log n/\xi^2)$ gives simultaneous relative accuracy $|\hat v_j-v_j|\leq\xi v_j$ over the $n$ input features with probability at least $1-\mathcal{O}(n^{-1})$, for $\xi\in(0,1)$.

Let $\mathcal{I}_{\mathrm{WIS},i}$ be the WIS-selected support in row $i$, and let $\mathcal{I}_i^*$ contain the $k$ largest expected gradient energies among all $n$ coordinates in that row. Set $\mathcal{A}_i=\mathcal{I}_{\mathrm{WIS},i}\cup\mathcal{I}_i^*$. For rows with nonzero gradient energy, suppose
\begin{equation}
w_{\min,i}=\min_{j\in\mathcal{A}_i}\mathbf{W}_{i,j}^2>0,
\qquad
w_{\max,i}=\max_{j\in\mathcal{A}_i}\mathbf{W}_{i,j}^2.
\nonumber
\end{equation}
Define
\begin{equation}
\kappa_i=\frac{\gamma w_{\min,i}}{\nu w_{\max,i}},
\qquad
\kappa_{i,\xi}=\kappa_i\frac{1-\xi}{1+\xi},
\qquad
\kappa_\xi=\min_{i:\omega_i>0}\kappa_{i,\xi}.
\nonumber
\end{equation}
These factors account for input--error dependence, weight variation within the compared supports, and calibration error.

\begin{lemma}[Gradient-Saliency Proxy Bound]\label{lemma:gradient_saliency}
Under Assumptions~\ref{assump:weak_data_corre} and~\ref{assump:sub_gaussian_act}, if $B=\Omega(K^2\log n/\xi^2)$, then with probability at least $1-\mathcal{O}(n^{-1})$,
\begin{equation}
C_{i,\mathrm{low}}\mathbf{S}_{i,j}^2
\leq\mathbb{E}[\mathbf{G}_{i,j}^2]
\leq C_{i,\mathrm{high}}\mathbf{S}_{i,j}^2
\nonumber
\end{equation}
simultaneously for $j\in\mathcal{A}_i$, where $C_{i,\mathrm{low}}/C_{i,\mathrm{high}}=\kappa_{i,\xi}$.
\end{lemma}

The lemma relates the Wanda ranking to expected gradient energy.

For a fixed support and nonzero expected gradient energy, define its energy capture ratio by
\begin{equation}
\rho\triangleq
\frac{\mathbb{E}[\|\mathbf{M}\odot\mathbf{G}\|_F^2]}
{\mathbb{E}[\|\mathbf{G}\|_F^2]}
=\sum_{i=1}^m\omega_i\rho_i,
\label{eq:global_energy_ratio_rho}
\end{equation}
where
\begin{equation}
\omega_i=
\frac{\mathbb{E}[\|\mathbf{G}_{i,:}\|_2^2]}{\mathbb{E}[\|\mathbf{G}\|_F^2]},
\qquad
\rho_i=
\frac{\mathbb{E}[\|\mathbf{M}_{i,:}\odot\mathbf{G}_{i,:}\|_2^2]}
{\mathbb{E}[\|\mathbf{G}_{i,:}\|_2^2]}.
\nonumber
\end{equation}
Rows with zero gradient energy contribute zero to the sum. Let $\rho^*=\sum_i\omega_i\rho_i^*$ be the capture ratio of the row-wise oracle supports $\mathcal{I}_i^*$.

\begin{theorem}[Energy Capture Relative to the Oracle]\label{theo:gradient_energy}
Under the conditions of Lemma~\ref{lemma:gradient_saliency}, with probability at least $1-\mathcal{O}(n^{-1})$,
\begin{equation}
\rho_{\mathrm{WIS}}
\geq\sum_i\omega_i\kappa_{i,\xi}\rho_i^*
\geq\kappa_\xi\rho^*.
\nonumber
\end{equation}
A uniform row-wise random mask of density $r$ has expected capture ratio $r$ at the same parameter value. Thus, $\kappa_\xi\rho^*>r$ is a sufficient condition for WIS to exceed this random-mask reference.
\end{theorem}

The result separates gradient-energy concentration, measured by $\rho^*$, from the accuracy of the saliency ranking, measured by $\kappa_\xi$. Calibration affects the latter through $\xi$.

\begin{figure*}[!th]
\centering
\begin{subfigure}[b]{0.32\linewidth}
\centering
\includegraphics[width=\linewidth]{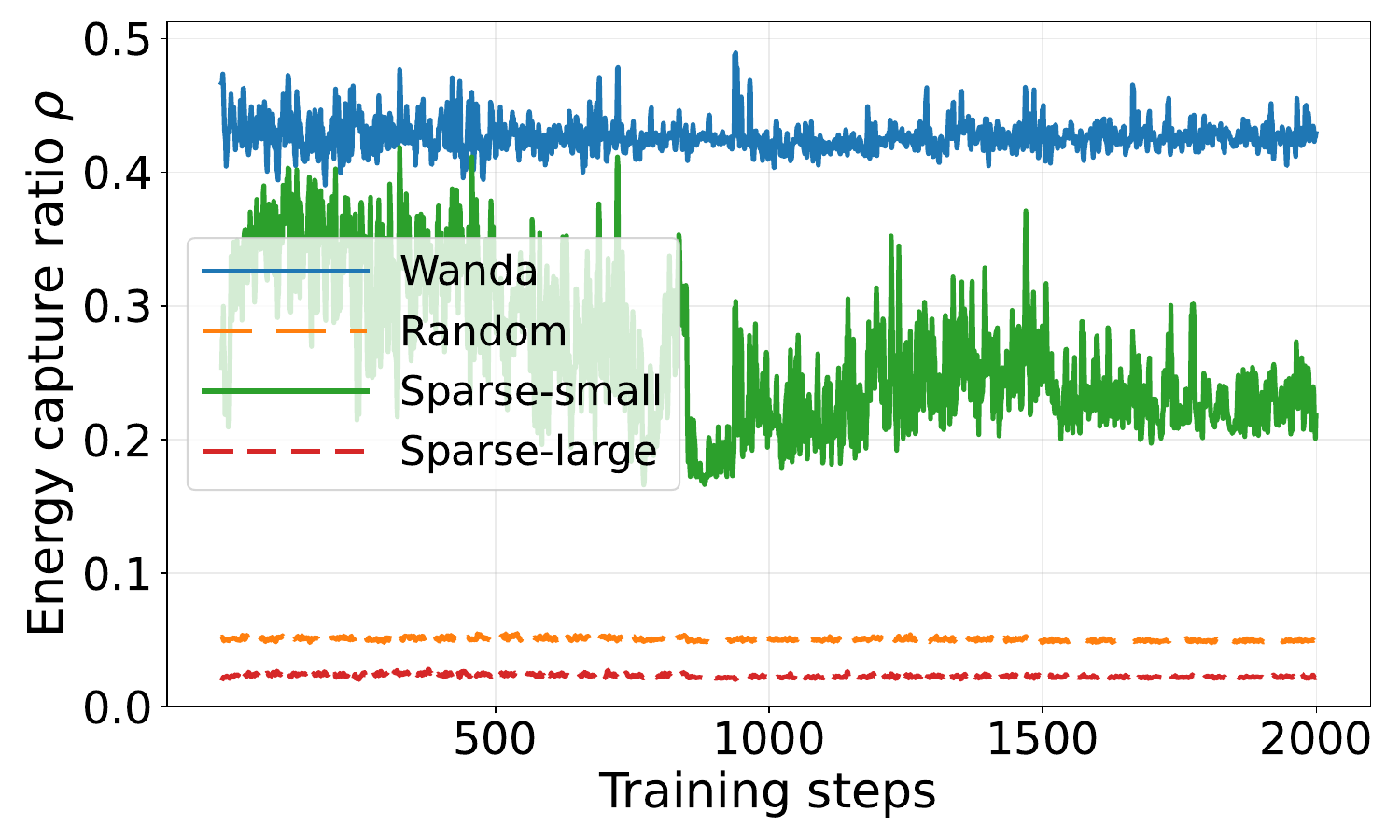}
\caption{RTE}
\end{subfigure}
\hfill
\begin{subfigure}[b]{0.32\linewidth}
\centering
\includegraphics[width=\linewidth]{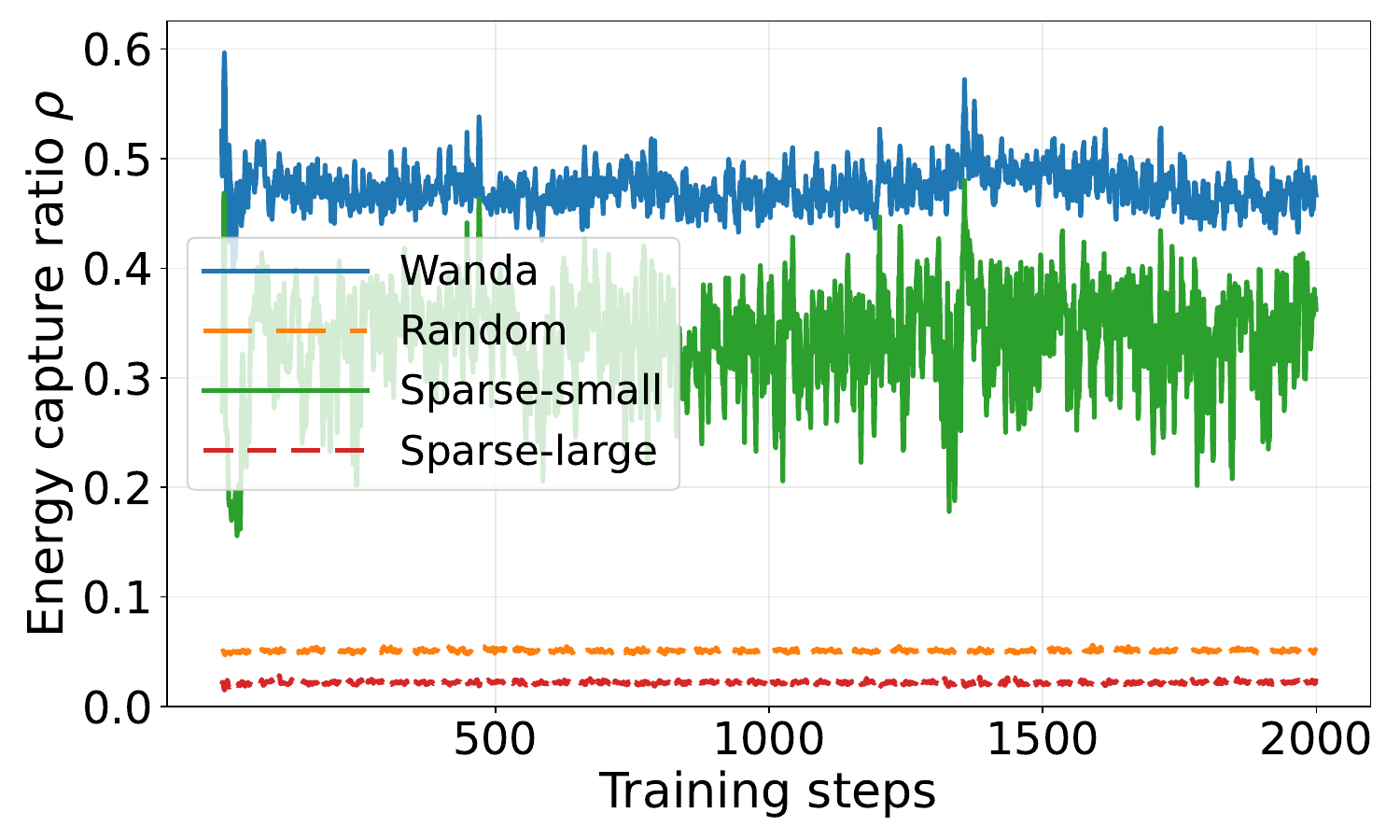}
\caption{QQP}
\end{subfigure}
\hfill
\begin{subfigure}[b]{0.32\linewidth}
\centering
\includegraphics[width=\linewidth]{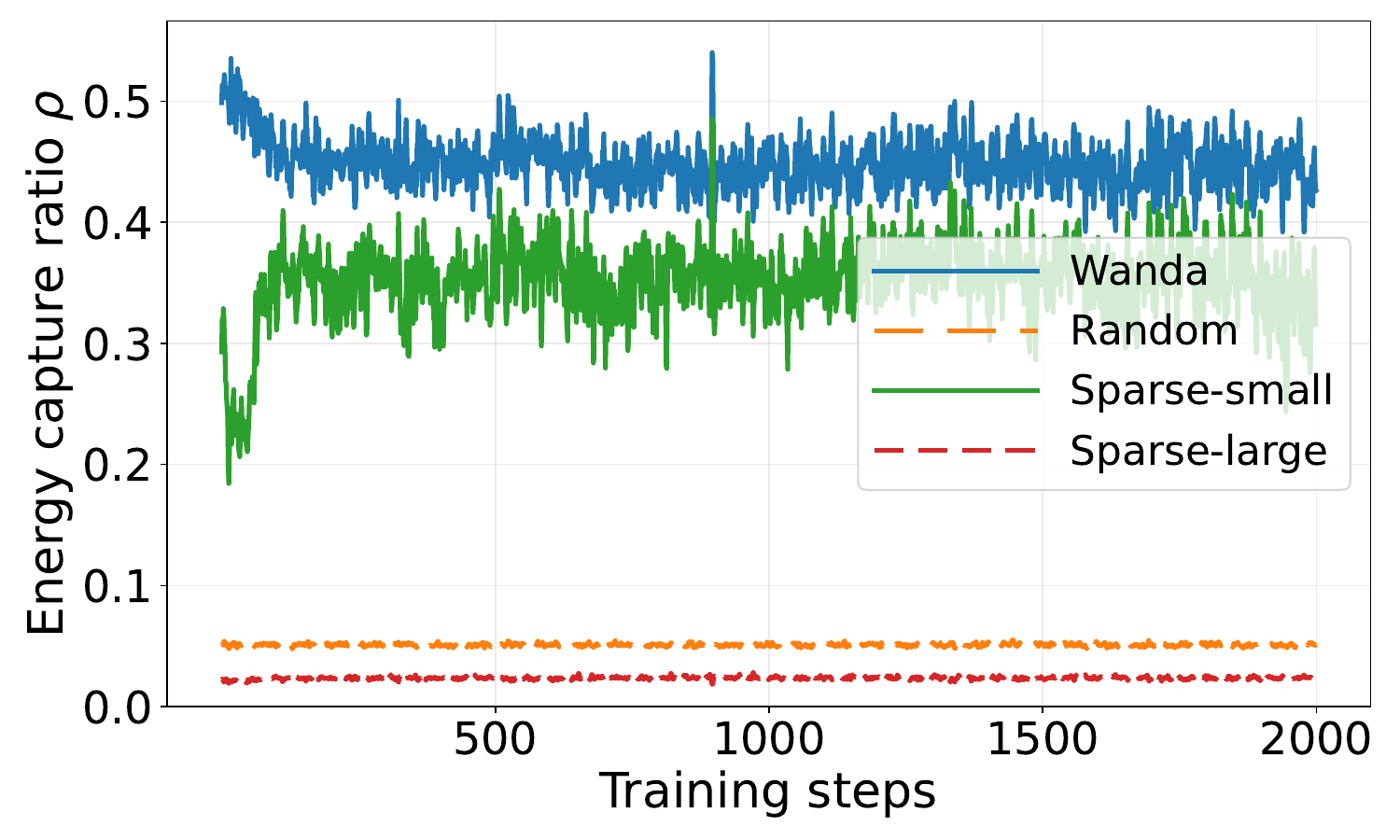}
\caption{MNLI}
\end{subfigure}
\caption{\textbf{Gradient energy capture ratio $\rho$ on LLaMA-3.2-3B at a density ratio of $r=0.05$.}
We compare uniform random masking, the small-magnitude mask used to represent the selection principle of Sparse-MeZO (W-small), the corresponding large-magnitude mask (W-large), and the Wanda-based WIS mask.}
\label{fig:energy_llama_glue_005}

\end{figure*}

Figure~\ref{fig:energy_llama_glue_005} compares support rules on common mini-batch gradient snapshots from an auxiliary AdamW run. All masks are constructed at step 0 and remain fixed. At $r=0.05$, the plotted WIS supports retain over $40\%$ of the gradient energy, compared with the $5\%$ random-mask expectation. WIS also captures more energy than W-small and W-large in these diagnostics. W-small exceeds random masking, while W-large falls below it, so the observed energy ordering is not determined by increasing weight magnitude. These measurements evaluate how the initial supports align with gradients encountered during the diagnostic run. Results at additional densities appear in Appendix~\ref{app:additional_energy_capture}.

We next consider estimator alignment at a fixed parameter value. For the mini-batch gradient $\mathbf{G}$, define $\rho$ by Eq.~\eqref{eq:global_energy_ratio_rho} using the mini-batch distribution. Let $s=d_{\mathrm{eff}}$ be the total number of selected coordinates and $\mathbf{U}=(\mathbf{Z}\odot\mathbf{M})/\sqrt r$, with standard Gaussian entries in $\mathbf{Z}$.

\begin{lemma}[Moment-Normalized Alignment]\label{lemma:cosine_sim}
For the directional-derivative estimator $\hat{\mathbf{G}}_0=\langle\mathbf{G},\mathbf{U}\rangle\mathbf{U}$ with $\rho>0$,
\begin{equation}
\operatorname{Align}(\hat{\mathbf{G}}_0,\mathbf{G})
\triangleq
\frac{\mathbb{E}[\langle\hat{\mathbf{G}}_0,\mathbf{G}\rangle]}
{\sqrt{\mathbb{E}[\|\hat{\mathbf{G}}_0\|_F^2]\mathbb{E}[\|\mathbf{G}\|_F^2]}}
=\sqrt{\frac{\rho}{s+2}}.
\nonumber
\end{equation}
At the same gradient distribution, averaging over uniform row-wise masks replaces $\rho$ by $r$.
\end{lemma}

The estimator $\hat{\mathbf{G}}_0$ is the $\epsilon\to0$ limit of the central-difference estimator. The lemma isolates the contribution of support quality at a fixed dimension. The next result includes the finite perturbation radius used by the algorithm.

\subsection{Convergence Analysis}

We analyze the fixed support produced at calibration. In this subsection, $\mathbf{W}$, $\mathbf{M}$, and $\mathbf{G}$ denote the full collections of parameter blocks, and squared norms sum over all blocks.

\begin{assumption}[$L$-Smooth Mini-Batch Losses]\label{assump:L_lipschitz}
Each mini-batch loss $f(\cdot,\mathcal{B})$ is differentiable and $L$-smooth. The population objective $f=\mathbb{E}_{\mathcal{B}}[f(\cdot,\mathcal{B})]$ is bounded below by $f^*$.
\end{assumption}

\begin{assumption}[Unbiased Gradients with Bounded Variance]\label{assump:bounded_variance}
Conditioned on the optimization history, $\mathbf{G}^t=\nabla f(\mathbf{W}^t,\mathcal{B}^t)$ satisfies
\begin{equation}
\mathbb{E}_t[\mathbf{G}^t]=\nabla f(\mathbf{W}^t),
\qquad
\mathbb{E}_t[\|\mathbf{G}^t-\nabla f(\mathbf{W}^t)\|_F^2]\leq\sigma^2.
\nonumber
\end{equation}
The perturbation is independent of the current mini-batch.
\end{assumption}

\begin{theorem}[Finite-Difference Convergence]\label{theo:convergence}
Suppose Assumptions~\ref{assump:L_lipschitz} and~\ref{assump:bounded_variance} hold. For a fixed mask, let $s=d_{\mathrm{eff}}$, $a=s+2$, $\Delta f=f(\mathbf{W}^0)-f^*$, and
\begin{equation}
\sigma_\epsilon^2
\triangleq\sigma^2+\frac{L^2\epsilon^2s(s+4)}{4r}.
\nonumber
\end{equation}
Assume a deterministic constant $\rho_{\min}\in(0,1]$ satisfies
\begin{equation}
\|\mathbf{M}\odot\nabla f(\mathbf{W}^t)\|_F^2
\geq\rho_{\min}\|\nabla f(\mathbf{W}^t)\|_F^2
\nonumber
\end{equation}
almost surely for $0\leq t<T$. With Gaussian perturbations, radius $\epsilon>0$, and $\eta_t=r/(4La\sqrt{t+1})$, the central-difference iterates satisfy
\begin{equation}
\begin{aligned}
\frac1T\sum_{t=0}^{T-1}\mathbb{E}[\|\nabla f(\mathbf{W}^t)\|_F^2]
&\leq\frac{8La\Delta f+\tfrac12\sigma_\epsilon^2(1+\log T)}{\rho_{\min}\sqrt T}
\\
&\quad+\frac{4L^2\epsilon^2s}{r\rho_{\min}}.
\end{aligned}
\nonumber
\end{equation}
\end{theorem}

The convergence bound depends on the selected dimension, the gradient energy retained during training, and the finite-difference radius. The result also highlights the role of support quality in sparse ZO optimization.

\begin{table*}[!th]
    \centering
    \caption{\textbf{Results on five GLUE tasks and SQuAD (five random seeds).} We report F1 scores for QQP, MRPC, and SQuAD, and accuracy (\%) for all other tasks ($\uparrow$). FT and LoRA denote full parameter fine-tuning and Low-Rank Adaptation, respectively, both utilizing the AdamW optimizer.}
    \label{tab:glu_qwen_llama_result}
    \resizebox{.9\linewidth}{!}{\begin{tabular}{l|l|cccccc|c}
    \toprule
       Models & Methods & \textbf{MNLI} & \textbf{RTE} & \textbf{QQP} & \textbf{MRPC} & \textbf{SST-2} & \textbf{SQuAD} & \textbf{Avg.} \\
       \midrule
       \multirow{8}{*}{LLaMA3.2-3B} 
       & Zero-shot & 38.3 & 49.4 & 53.7 & 78.3 & 58.8 & 55.0 & 55.6 \\
       \cmidrule(){2-9}
       & FT & 82.6$_{\pm 0.6}$ & 87.3$_{\pm 0.8}$ & 79.2$_{\pm 1.0}$ & 86.9$_{\pm 1.0}$ & 95.1$_{\pm 0.5}$ & 88.2$_{\pm 0.4}$ & 86.5 \\
       & LoRA & 83.4$_{\pm 0.7}$ & 88.8$_{\pm 0.7}$ & 79.2$_{\pm 1.0}$ & 85.8$_{\pm 1.3}$ & 94.8$_{\pm 0.1}$ & 88.0$_{\pm 0.0}$ & 86.6 \\
       \cmidrule(){2-9}
       & MeZO & 49.4$_{\pm 3.0}$ & 70.3$_{\pm 2.1}$ & 65.0$_{\pm 2.1}$ & 81.4$_{\pm 0.6}$ & 93.9$_{\pm 0.8}$ & 86.7$_{\pm 0.6}$ & 74.4 \\
       & Sparse-MeZO & 56.7$_{\pm 2.6}$ & 77.6$_{\pm 3.8}$ & 70.6$_{\pm 0.6}$ & 81.8$_{\pm 0.9}$ & 93.6$_{\pm 0.3}$ & 86.4$_{\pm 0.7}$ & 77.7 \\
       & Random-MeZO & 52.1$_{\pm 1.8}$ & 74.3$_{\pm 0.7}$ & 67.4$_{\pm 1.3}$ & 82.6$_{\pm 1.1}$ & 93.9$_{\pm 0.4}$ & 88.1$_{\pm 0.6}$ & 76.4 \\
       \cmidrule(){2-9}
       \rowcolor{blue4}\cellcolor{white} & WIS-MeZO & 61.2$_{\pm 1.1}$ & {78.3}$_{\pm 1.3}$ & 68.7$_{\pm 1.0}$ & {\textbf{83.1}}$_{\pm 1.1}$ & \textbf{94.0}$_{\pm 0.2}$ & 89.2$_{\pm 0.8}$ & 79.1 \\
       \rowcolor{blue2}\cellcolor{white} & WISD-MeZO & \textbf{61.4}$_{\pm 1.3}$ & \textbf{80.1}$_{\pm 2.2}$ & \textbf{71.2}$_{\pm 0.9}$ & 82.9$_{\pm 1.0}$ & \textbf{94.0}$_{\pm 0.5}$ & \textbf{89.4}$_{\pm 1.6}$ & \textbf{79.8} \\
       \midrule 
       \multirow{8}{*}{Qwen2.5-3B} 
       & Zero-shot & 74.8 & 80.5 & 34.1 & 51.1 & 56.0 & 83.6 & 63.4 \\
       \cmidrule(){2-9}
       & FT & 86.9$_{\pm 0.5}$ & 87.8$_{\pm 0.6}$ & 81.2$_{\pm 0.1}$ & 89.1$_{\pm 0.8}$ & 94.1$_{\pm 0.4}$ & 85.6$_{\pm 0.2}$ & 87.5 \\
       & LoRA & 86.6$_{\pm 0.4}$ & 87.9$_{\pm 0.4}$ & 79.5$_{\pm 0.2}$ & 89.6$_{\pm 1.0}$ & 93.6$_{\pm 0.4}$ & 87.2$_{\pm 1.0}$ & 87.4 \\
       \cmidrule(){2-9}
       & MeZO & 84.8$_{\pm 0.7}$ & 84.8$_{\pm 0.9}$ & 78.7$_{\pm 1.1}$ & 84.9$_{\pm 1.8}$ & 92.6$_{\pm 0.2}$ & 86.6$_{\pm 0.9}$ & 85.4 \\
       & Sparse-MeZO & 86.1$_{\pm 1.0}$ & 86.6$_{\pm 1.5}$ & 81.3$_{\pm 0.8}$ & 85.9$_{\pm 1.0}$ & 93.6$_{\pm 0.6}$ & 86.3$_{\pm 0.7}$ & 86.6 \\
       & Random-MeZO & 85.1$_{\pm 1.3}$ & 85.2$_{\pm 1.6}$ & 81.3$_{\pm 0.6}$ & 84.9$_{\pm 1.1}$ & 93.1$_{\pm 0.2}$ & 86.6$_{\pm 0.6}$ & 86.0 \\
       \cmidrule(){2-9}
        \rowcolor{blue4}\cellcolor{white} & WIS-MeZO & \textbf{86.7}$_{\pm 0.4}$ & \textbf{87.0}$_{\pm 0.7}$ & 81.5$_{\pm 0.6}$ & 85.6$_{\pm 1.1}$ & 93.6$_{\pm 0.2}$ & \textbf{87.1}$_{\pm 0.8}$ & 86.9 \\
        \rowcolor{blue2}\cellcolor{white} & WISD-MeZO & 86.3$_{\pm 0.9}$ & \textbf{87.0}$_{\pm 1.0}$ & \textbf{81.6}$_{\pm 0.7}$ & \textbf{86.4}$_{\pm 1.2}$ & \textbf{93.7}$_{\pm 0.1}$ & 86.8$_{\pm 1.1}$ & \textbf{87.0} \\
       \bottomrule
    \end{tabular}
    }
\end{table*}

\section{Experiments}

We now examine how task-conditioned support selection affects downstream fine-tuning. We evaluate five GLUE tasks~\cite{wang2018glue}, SQuAD, and five SuperGLUE tasks~\cite{wang2019superglue}, followed by commonsense and question-answering benchmarks across larger model scales. Additional efficiency and sensitivity results appear in the appendix.

\subsection{Fine-Tuning on GLUE, SQuAD, and SuperGLUE}

We evaluate WIS-MeZO and its dynamic variant WISD-MeZO on LLaMA-3.2-3B~\cite{grattafiori2024llama3} and Qwen2.5-3B~\cite{qwen2025qwen25technicalreport}. All ZO methods use a batch size of 16 and train for 20,000 steps. Hyperparameters are reported in Appendix~\ref{app:experimental_details}.

The first-order baselines are full-parameter fine-tuning (FT) with AdamW~\cite{Kingma2014_adam,loshchilov2018decoupled_adamw} and LoRA~\cite{hu2022lora}. The ZO baselines are MeZO~\cite{malladi2023_mezo}, Sparse-MeZO~\cite{liu2025sparse}, which selects small-magnitude weights, and Random-MeZO, which uses random sparse supports. These comparisons evaluate the effect of coordinate selection within sparse ZO fine-tuning.

Tables~\ref{tab:glu_qwen_llama_result} and~\ref{tab:super_glue_qwen_llama_result} report the task results. Figure~\ref{fig:learning_curve_optimizer} compares validation loss against optimization steps over five random seeds.

\begin{table*}[!th]
    \centering
    \caption{\textbf{Experimental results on SuperGLUE.} WIS-based methods achieve the strongest aggregate ZO performance on both model families.}
    \label{tab:super_glue_qwen_llama_result}
    \resizebox{.75\linewidth}{!}{
    \begin{tabular}{l|l|ccccc|c}
    \toprule
       Models & Methods & \textbf{CB} & \textbf{MultiRC} & \textbf{WSC} & \textbf{ReCoRD} & \textbf{BoolQ} & \textbf{Avg.} \\
       \midrule
       \multirow{8}{*}{LLaMA3.2-3B} & Zero-shot & 44.6 & 47.0 & 36.5 & 79.7 & 61.6 & 53.9 \\
       \cmidrule(){2-8}
       & FT & 91.0 & 83.6 & 57.6 & 81.6 & 83.1 & 79.4 \\
       & LoRA & 92.8 & 85.2 & 62.5 & 81.4 & 84.3 & 81.2 \\
       \cmidrule(){2-8}
       & MeZO & 71.4 & 72.4 & 61.5 & 81.2 & 77.6 & 72.8 \\
       & Sparse-MeZO & 67.8 & \textbf{79.7} & 62.5 & 81.4 & 81.3 & 74.5 \\
       & Random-MeZO & 75.0 & 74.2 & 61.5 & 81.0 & 78.9 & 74.1 \\
       \cmidrule(){2-8}
       \rowcolor{blue4}\cellcolor{white}& WIS-MeZO & \textbf{82.1} & 76.3 & 63.4 & \textbf{82.0} & {82.6} & 77.3 \\
       \rowcolor{blue2}\cellcolor{white}& WISD-MeZO & \textbf{82.1} & 76.3 & \textbf{64.4} & 81.9 & \textbf{83.0} & \textbf{77.6} \\
       \midrule 
       \multirow{8}{*}{Qwen2.5-3B} & Zero-shot & 30.3 & 65.8 & 63.4 & 72.1 & 70.3 & 60.3 \\
       \cmidrule(){2-8}
       & FT & 96.4 & 88.0 & 68.4 & 74.4 & 80.9 & 81.6\\
       & LoRA & 94.6 & 86.9 & 67.4 & 76.0 & 85.1 & 82.0 \\
       \cmidrule(){2-8}
       & MeZO & 96.4 & 83.2 & 66.3 & 73.7 & 80.9 & 80.1 \\
       & Sparse-MeZO & 92.8 & 86.1 & 67.3 & \textbf{75.6} & 80.1 & 80.3 \\
       & Random-MeZO & 96.4 & 85.6 & 65.3 & 74.5 & 81.6 & 80.9 \\
       \cmidrule(){2-8}
       \rowcolor{blue4}\cellcolor{white} & WIS-MeZO & 96.4 & 86.4 & 68.2 & 74.9 & {82.6} & 81.7 \\
       \rowcolor{blue2}\cellcolor{white} & WISD-MeZO & \textbf{98.2} & \textbf{86.9} & \textbf{71.1} & \textbf{75.6} & \textbf{82.8} & \textbf{82.9} \\
       \bottomrule
    \end{tabular}
    }
\end{table*}

WIS-MeZO improves the average score over MeZO, Random-MeZO, and Sparse-MeZO in all four model--benchmark combinations. On LLaMA-3.2-3B, its average scores are 79.1 on the GLUE/SQuAD tasks and 77.3 on SuperGLUE, compared with 77.7 and 74.5 for Sparse-MeZO. The corresponding Qwen2.5-3B scores are 86.9 and 81.7, compared with 86.6 and 80.3. The size of the gain varies across backbones and tasks, with the larger improvements occurring on LLaMA-3.2-3B.

Beyond final task scores, the validation-loss curves in Figure~\ref{fig:learning_curve_optimizer} show that WIS-MeZO reaches comparable loss levels in fewer optimization steps than MeZO on RTE, MNLI, and BoolQ. This improvement is consistent with the gradient-energy results in Section~\ref{sec:theoretical_analysis}. Runtime and memory measurements are reported in Section~\ref{sec:memory_speed}.

Periodic recalibration in WISD-MeZO increases the average score in all four model--benchmark combinations. Fixed-mask WIS-MeZO is the primary method, and WISD-MeZO characterizes the effect of refreshing the support during training.

\subsection{Scaling to Large-Scale LLMs}

Having compared support-selection strategies on 3B models, we next examine whether the improvement over MeZO extends to larger models. We evaluate fixed-mask WIS-MeZO on Qwen2.5 models from 3B to 32B parameters using ARC-E, ARC-C~\cite{clark2018arc}, HellaSwag~\cite{zellers2019hellaswag}, WinoGrande~\cite{sakaguchi2021winogrande}, and OpenBookQA~\cite{mihaylov2018openbook}. Table~\ref{tab:qwen_3b-32b_results} shows higher average scores at every scale and performance matching or exceeding MeZO on every task. On WinoGrande, WIS-MeZO also improves over the zero-shot model at all four scales.

\begin{table*}[!ht]
    \centering
    \caption{\textbf{Experimental results for Qwen2.5-3B to 32B.} WIS-MeZO consistently improves average performance over MeZO across all model scales.}
    \label{tab:qwen_3b-32b_results}
    \resizebox{.9\linewidth}{!}{
    \begin{tabular}{l|l|ccccc|c}
    \toprule
     Models & Methods & ARC-E & ARC-C & HellaSwag & WinoGrande & OpenBookQA & Avg. \\
      \midrule
     \multirow{3}{*}{Qwen2.5-3B} & Zero-shot & 70.0 & 48.4 & 68.0 & 66.3 & 40.8 & 58.7 \\
     & MeZO & 82.9 & 57.8 & 69.9 & 66.3 & 49.0 & 65.1 \\
     \rowcolor{blue4}\cellcolor{white}& WIS-MeZO & \textbf{85.4} & \textbf{60.8} & \textbf{74.4} & \textbf{69.7} & \textbf{54.0} & \textbf{68.8} \\
     \midrule
     \multirow{3}{*}{Qwen2.5-7B} & Zero-shot & 74.9 & 52.5 & 73.8 & 68.9 & 43.8 & 62.7 \\
     & MeZO & 83.0 & 59.2 & 74.9 & 66.5 & 51.2 & 66.9 \\
     \rowcolor{blue4}\cellcolor{white}& WIS-MeZO & \textbf{85.0} & \textbf{61.2} & \textbf{79.4} & \textbf{70.7} & \textbf{53.8} & \textbf{70.0} \\
     \midrule
     \multirow{3}{*}{Qwen2.5-14B} & Zero-shot & 75.4 & 53.8 & 75.8 & 70.6 & 43.2 & 63.7 \\
     & MeZO & 84.0 & 67.2 & 77.2 & 73.8 & 57.2 & 71.8 \\
     \rowcolor{blue4}\cellcolor{white}& WIS-MeZO & \textbf{87.8} & \textbf{70.9} & \textbf{81.5} & \textbf{76.6} & \textbf{58.8} & \textbf{75.1}\\
     \midrule
     \multirow{3}{*}{Qwen2.5-32B} & Zero-shot & 74.5 & 54.5 & 79.0 & 74.3 & 43.6 & 65.1\\
     & MeZO & 85.9 & \textbf{69.8} & 78.7 & 74.9 & 52.4 & 72.3 \\
     \rowcolor{blue4}\cellcolor{white}& WIS-MeZO & \textbf{88.7} & \textbf{69.8} & \textbf{84.5} & \textbf{77.9} & \textbf{56.2} & \textbf{75.4}\\
     \bottomrule
    \end{tabular}
    }
\end{table*}

\section{Additional Experimental Results}

\begin{table}[!ht]
    \centering
    \caption{Peak GPU memory (GB, batch size=16) and iteration speed (iteration/s) for fine-tuning Qwen2.5-3B.}
    \label{tab:mem_speed_qwen25-3b}
    \begin{tabular}{l|cc|cc|cc}
    \toprule
      \multirow{2}{*}{Methods} & \multicolumn{2}{c}{\textbf{SST-2}} & \multicolumn{2}{c}{\textbf{ReCoRD}} & \multicolumn{2}{c
      }{\textbf{CB}} \\
      & Mem. & Iter/s & Mem. & Iter/s & Mem. & Iter/s \\
      \midrule
      FT     & 35.5 & 3.37 & 87.4 & 0.92 & 94.5 & 0.75 \\
      LoRA   & 22.1 & 4.73 & 73.5 & 1.33 & 85.0 & 0.94 \\
      MeZO   & 9.70 & 3.50 & 21.5 & 1.39 & 45.1 & 0.99 \\
      Sparse-MeZO & 16.2 & 2.98 & 27.3 & 1.30 & 51.0 & 0.95 \\
      \rowcolor{blue4}WIS-MeZO & 11.6 & 3.16 & 25.1 & 1.33 & 47.0 & 0.97 \\
      \bottomrule
    \end{tabular}
\end{table}

\begin{table}[!ht]
    \centering
    \caption{\textbf{End-to-end wall-clock time (hours) for Qwen2.5-3B under the common ZO training budget.}}
    \label{tab:wallclock_qwen25-3b}
    \begin{tabular}{l|c|c|c}
    \toprule
      {Methods} & {\textbf{SST-2}} & {\textbf{ReCoRD}} & {\textbf{CB}} \\
      \midrule
      MeZO        & 1.59 & 4.00 & 5.61 \\
      Sparse-MeZO & 1.86 & 4.27 & 5.85 \\
      \rowcolor{blue4}WIS-MeZO & 1.76 & 4.18 & 5.73 \\
      \bottomrule
    \end{tabular}
\end{table}

\subsection{Memory and Speed Comparison}
\label{sec:memory_speed}

We report peak GPU memory, iteration throughput, and total training time in Tables~\ref{tab:mem_speed_qwen25-3b} and~\ref{tab:wallclock_qwen25-3b}. Table~\ref{tab:mem_speed_qwen25-32b} compares standard and bit-packed masks on Qwen2.5-32B.

On the 3B model, WIS-MeZO uses less peak memory and has higher throughput than the evaluated Sparse-MeZO implementation. Its wall-clock times on SST-2, ReCoRD, and CB are 1.76, 4.18, and 5.73 hours. These are 2.0\%--5.4\% lower than Sparse-MeZO and 2.1\%--10.7\% higher than MeZO. The comparison quantifies the computational cost of masked updates alongside the validation-loss curves indexed by optimization step.

At 32B, a 1-bit mask reduces WIS-MeZO's memory overhead relative to MeZO to below 3GB on both tasks while keeping similar throughput. This compact representation supports larger-scale sparse ZO fine-tuning.

\begin{table}[!ht]
    \centering
    \caption{Peak GPU memory (GB, batch size=16) and iteration speed (iteration/s) for fine-tuning Qwen2.5-32B.}
    \label{tab:mem_speed_qwen25-32b}
    \resizebox{0.5\linewidth}{!}{\begin{tabular}{l|cc|cc}
    \toprule
        \multirow{2}{*}{Methods} & \multicolumn{2}{c}{\textbf{ARC-C}} & \multicolumn{2}{c}{\textbf{HellaSwag}}  \\
        & Mem. & Iter/s & Mem. & Iter/s \\
        \midrule
        FT & 600.5 & 0.126 & 664.8 & 0.074 \\
        LoRA & 256.0 & 0.161 & 282.2 & 0.105 \\
        MeZO & 97.29 & 0.174 & 104.8 & 0.104  \\
        \rowcolor{blue4}WIS-MeZO & 122.0 & 0.169 & 132.3 & 0.102  \\
        \rowcolor{blue2}WIS-MeZO (1-bit) & 98.84 & 0.160 & 107.7 & 0.099 \\
        \bottomrule
    \end{tabular}
    }
\end{table}

\subsection{Study on Density Ratio-$r$}
The density ratio $r$ sets the number of coordinates available for perturbation. Table~\ref{tab:qwen25_3b_ablation_r} compares the evaluated densities on Qwen2.5-3B.

\begin{table}[!ht]
    \centering
    \caption{The effects of density ratio-$r$ for Fine-tuning Qwen2.5-3B with WIS-MeZO.}
    \label{tab:qwen25_3b_ablation_r}
    \resizebox{0.6\linewidth}{!}{
    \begin{tabular}{l|ccccc}
    \toprule
       Ratio-$r$ & ARC-E & ARC-C & HellaSwag & WinoGrande & OpenBookQA \\
       \midrule
       0.05 & 85.0 & 57.8 & 71.8 & 68.6 & \textbf{54.0} \\
       0.1  & \textbf{85.4} & \textbf{60.8} & \textbf{74.4} & \textbf{69.7} & 52.8 \\
       0.2  & 83.8 & 59.5 & 72.0 & 69.5 & 53.6 \\
       0.4  & 84.2 & 59.8 & 71.4 & 67.1 & 52.8 \\
       \midrule
       MeZO & 82.9 & 57.8 & 69.9 & 66.3 & 49.0 \\
       \bottomrule
    \end{tabular}
    }
\end{table}

WIS-MeZO matches or improves on MeZO at every reported task--density combination. A density of $r=0.1$ gives the highest score on four of the five tasks in this table, while OpenBookQA performs best at $r=0.05$.

\subsection{Effects of Calibration Size and Computation Cost}
\label{sec:calibration_size}

Table~\ref{tab:qwen25_3b_ablation_samples} studies calibration-set size on Qwen2.5-3B. Performance improves with larger calibration sets and stabilizes around 128 examples, which we use by default. Table~\ref{tab:qwen25_3b_ablation_samples_time} reports the total calibration time. With 128 examples, mask construction takes 0.53--1.26 seconds and is performed once for fixed-mask WIS-MeZO.

\begin{table}[!ht]
\centering
\caption{Effects of the calibration size on WIS-MeZO fine-tuning of
Qwen2.5-3B.}
\label{tab:qwen25_3b_ablation_samples}
\resizebox{0.7\linewidth}{!}{
\begin{tabular}{l|ccccc}
\toprule
Calibration size & ARC-E & ARC-C & HellaSwag & WinoGrande & OpenBookQA \\
\midrule
32  & 85.0 & 60.5 & 74.0 & 67.4 & 54.0 \\
64  & 85.2 & 60.5 & 74.3 & 67.8 & 53.0 \\
128 (default) & 85.4 & 60.8 & 74.4 & 69.7 & 54.0 \\
256 & 85.4 & 60.5 & 74.6 & 69.7 & 54.0 \\
512 & 85.9 & 60.8 & 74.5 & 69.7 & 53.9 \\
\bottomrule
\end{tabular}
}
\end{table}

\begin{table}[!ht]
\centering
\caption{Total calibration time (seconds) for Qwen2.5-3B.}
\label{tab:qwen25_3b_ablation_samples_time}
\resizebox{0.7\linewidth}{!}{
\begin{tabular}{l|ccccc}
\toprule
Calibration size & ARC-E & ARC-C & HellaSwag & WinoGrande & OpenBookQA \\
\midrule
32  & 0.32 & 0.33 & 0.43 & 0.26 & 0.27 \\
64  & 0.49 & 0.49 & 0.72 & 0.34 & 0.38 \\
128 (default) & 0.80 & 0.89 & 1.26 & 0.53 & 0.60 \\
256 & 1.49 & 1.55 & 2.36 & 0.91 & 1.00 \\
512 & 2.70 & 2.98 & 4.59 & 1.66 & 1.84 \\
\bottomrule
\end{tabular}
}
\end{table}

\section{Conclusion}

We present WIS-MeZO, which uses downstream Wanda scores to select sparse perturbation coordinates for memory-efficient LLM fine-tuning. Our gradient-energy analysis explains the importance of support selection, while experiments across 16 benchmarks and models from 3B to 32B show improved aggregate performance over MeZO. By combining forward-only updates with one-time calibration and compact mask storage, WIS-MeZO also offers a promising direction for resource-constrained on-device adaptation.

\bibliographystyle{plainnat}
\bibliography{reference}

@article{malladi2023_mezo,
  title={Fine-tuning language models with just forward passes},
  author={Malladi, Sadhika and Gao, Tianyu and Nichani, Eshaan and Damian, Alex and Lee, Jason D and Chen, Danqi and Arora, Sanjeev},
  journal={Advances in Neural Information Processing Systems},
  volume={36},
  pages={53038--53075},
  year={2023}
}

@inproceedings{
wanda,
title={A Simple and Effective Pruning Approach for Large Language Models},
author={Mingjie Sun and Zhuang Liu and Anna Bair and J Zico Kolter},
booktitle={The Twelfth International Conference on Learning Representations},
year={2024}
}

@article{dettmers2022gpt3_int8,
  title={Gpt3. int8 (): 8-bit matrix multiplication for transformers at scale},
  author={Dettmers, Tim and Lewis, Mike and Belkada, Younes and Zettlemoyer, Luke},
  journal={Advances in neural information processing systems},
  volume={35},
  pages={30318--30332},
  year={2022}
}

@article{spall2002multivariate,
  title={Multivariate stochastic approximation using a simultaneous perturbation gradient approximation},
  author={Spall, James C},
  journal={IEEE transactions on automatic control},
  volume={37},
  number={3},
  pages={332--341},
  year={2002},
  publisher={IEEE}
}

@inproceedings{ilyas2018black,
  title={Black-box adversarial attacks with limited queries and information},
  author={Ilyas, Andrew and Engstrom, Logan and Athalye, Anish and Lin, Jessy},
  booktitle={International conference on machine learning},
  pages={2137--2146},
  year={2018},
  organization={PMLR}
}

@article{zo_sgd,
  title={Stochastic first-and zeroth-order methods for nonconvex stochastic programming},
  author={Ghadimi, Saeed and Lan, Guanghui},
  journal={SIAM journal on optimization},
  volume={23},
  number={4},
  pages={2341--2368},
  year={2013},
  publisher={SIAM}
}

@article{chen2019zo_adamm,
  title={Zo-adamm: Zeroth-order adaptive momentum method for black-box optimization},
  author={Chen, Xiangyi and Liu, Sijia and Xu, Kaidi and Li, Xingguo and Lin, Xue and Hong, Mingyi and Cox, David},
  journal={Advances in neural information processing systems},
  volume={32},
  year={2019}
}

@inproceedings{
chen2025lozo,
title={Enhancing Zeroth-order Fine-tuning for Language Models with Low-rank Structures},
author={Yiming Chen and Yuan Zhang and Liyuan Cao and Kun Yuan and Zaiwen Wen},
booktitle={The Thirteenth International Conference on Learning Representations},
year={2025}
}

@inproceedings{zhao2025pazo,
  title={PaZO: Preconditioned Accelerated Zeroth-Order Optimization for Fine-Tuning LLMs},
  author={Zhao, Hanzhen and Ding, Shihong and Fang, Cong and Lin, Zhouchen},
  booktitle={The Thirty-ninth Annual Conference on Neural Information Processing Systems},
  year={2025},
}

@inproceedings{
liu2025sparse,
title={Sparse Me{ZO}: Less Parameters for Better Performance in Zeroth-Order {LLM} Fine-Tuning},
author={Yong Liu and Zirui Zhu and Chaoyu Gong and Minhao Cheng and Cho-Jui Hsieh and Yang You},
booktitle={The Thirty-ninth Annual Conference on Neural Information Processing Systems},
year={2025}
}

@inproceedings{
dang2026fzoo,
title={{FZOO}: Fast Zeroth-Order Optimizer for Fine\nobreakdash-Tuning Large Language Models towards Adam\nobreakdash-Scale Speed},
author={Sizhe Dang and yangyangGuo and Yanjun Zhao and Haishan Ye and xiaodong zheng and Guang Dai and Ivor Tsang},
booktitle={The Fourteenth International Conference on Learning Representations},
year={2026}
}

@inproceedings{yu2025subzero,
  title={Zeroth-order fine-tuning of llms in random subspaces},
  author={Yu, Ziming and Zhou, Pan and Wang, Sike and Li, Jia and Tian, Mi and Huang, Hua},
  booktitle={Proceedings of the IEEE/CVF International Conference on Computer Vision},
  pages={4475--4485},
  year={2025}
}

@inproceedings{frantar2023sparsegpt,
  title={Sparsegpt: Massive language models can be accurately pruned in one-shot},
  author={Frantar, Elias and Alistarh, Dan},
  booktitle={International conference on machine learning},
  pages={10323--10337},
  year={2023},
  organization={PMLR}
}

@inproceedings{evci2020rigging,
  title={Rigging the lottery: Making all tickets winners},
  author={Evci, Utku and Gale, Trevor and Menick, Jacob and Castro, Pablo Samuel and Elsen, Erich},
  booktitle={International conference on machine learning},
  pages={2943--2952},
  year={2020},
  organization={PMLR}
}

@article{joo2026surprising,
  title={On Surprising Effectiveness of Masking Updates in Adaptive Optimizers},
  author={Joo, Taejong and Xia, Wenhan and Kim, Cheolmin and Zhang, Ming and Ie, Eugene},
  journal={arXiv preprint arXiv:2602.15322},
  year={2026}
}

@article{chen2023deepzero,
  title={Deepzero: Scaling up zeroth-order optimization for deep model training},
  author={Chen, Aochuan and Zhang, Yimeng and Jia, Jinghan and Diffenderfer, James and Liu, Jiancheng and Parasyris, Konstantinos and Zhang, Yihua and Zhang, Zheng and Kailkhura, Bhavya and Liu, Sijia},
  journal={arXiv preprint arXiv:2310.02025},
  year={2023}
}

@article{cai2022zeroth,
  title={Zeroth-order regularized optimization (zoro): Approximately sparse gradients and adaptive sampling},
  author={Cai, HanQin and McKenzie, Daniel and Yin, Wotao and Zhang, Zhenliang},
  journal={SIAM Journal on Optimization},
  volume={32},
  number={2},
  pages={687--714},
  year={2022},
  publisher={SIAM}
}

@inproceedings{ohta2020sparse,
  title={Sparse perturbations for improved convergence in stochastic zeroth-order optimization},
  author={Ohta, Mayumi and Berger, Nathaniel and Sokolov, Artem and Riezler, Stefan},
  booktitle={International Conference on Machine Learning, Optimization, and Data Science},
  pages={39--64},
  year={2020},
  organization={Springer}
}

@inproceedings{gu2021efficient,
  title={Efficient on-chip learning for optical neural networks through power-aware sparse zeroth-order optimization},
  author={Gu, Jiaqi and Feng, Chenghao and Zhao, Zheng and Ying, Zhoufeng and Chen, Ray T and Pan, David Z},
  booktitle={Proceedings of the AAAI conference on artificial intelligence},
  volume={35},
  number={9},
  pages={7583--7591},
  year={2021}
}

@inproceedings{wang2018glue,
  title={GLUE: A multi-task benchmark and analysis platform for natural language understanding},
  author={Wang, Alex and Singh, Amanpreet and Michael, Julian and Hill, Felix and Levy, Omer and Bowman, Samuel},
  booktitle={Proceedings of the 2018 EMNLP workshop BlackboxNLP: Analyzing and interpreting neural networks for NLP},
  pages={353--355},
  year={2018}
}

@article{wang2019superglue,
  title={Superglue: A stickier benchmark for general-purpose language understanding systems},
  author={Wang, Alex and Pruksachatkun, Yada and Nangia, Nikita and Singh, Amanpreet and Michael, Julian and Hill, Felix and Levy, Omer and Bowman, Samuel},
  journal={Advances in neural information processing systems},
  volume={32},
  year={2019}
}

@inproceedings{
loshchilov2018decoupled_adamw,
title={Decoupled Weight Decay Regularization},
author={Ilya Loshchilov and Frank Hutter},
booktitle={International Conference on Learning Representations},
year={2019}
}

@inproceedings{Kingma2014_adam,
  author       = {Diederik P. Kingma and
                  Jimmy Ba},
  title        = {Adam: {A} Method for Stochastic Optimization},
  booktitle={International Conference on Learning Representations},
  year         = {2015},
}

@inproceedings{
hu2022lora,
title={Lo{RA}: Low-Rank Adaptation of Large Language Models},
author={Edward J Hu and yelong shen and Phillip Wallis and Zeyuan Allen-Zhu and Yuanzhi Li and Shean Wang and Lu Wang and Weizhu Chen},
booktitle={International Conference on Learning Representations},
year={2022}
}

@article{qwen2025qwen25technicalreport,
    title   = {Qwen2.5 Technical Report}, 
    author  = {An Yang and Baosong Yang and Beichen Zhang and Binyuan Hui and Bo Zheng and Bowen Yu and Chengyuan Li and Dayiheng Liu and Fei Huang and Haoran Wei and Huan Lin and Jian Yang and Jianhong Tu and Jianwei Zhang and Jianxin Yang and Jiaxi Yang and Jingren Zhou and Junyang Lin and Kai Dang and Keming Lu and Keqin Bao and Kexin Yang and Le Yu and Mei Li and Mingfeng Xue and Pei Zhang and Qin Zhu and Rui Men and Runji Lin and Tianhao Li and Tingyu Xia and Xingzhang Ren and Xuancheng Ren and Yang Fan and Yang Su and Yichang Zhang and Yu Wan and Yuqiong Liu and Zeyu Cui and Zhenru Zhang and Zihan Qiu},
    journal = {arXiv preprint arXiv:2412.15115},
    year    = {2024}
}

@article{grattafiori2024llama3,
  title={The llama 3 herd of models},
  author={Grattafiori, Aaron and Dubey, Abhimanyu and Jauhri, Abhinav and Pandey, Abhinav and Kadian, Abhishek and Al-Dahle, Ahmad and Letman, Aiesha and Mathur, Akhil and Schelten, Alan and Vaughan, Alex and others},
  journal={arXiv preprint arXiv:2407.21783},
  year={2024}
}

@inproceedings{
guo2025zerothorder_szoq,
title={Zeroth-Order Fine-Tuning of {LLM}s with Transferable Static Sparsity},
author={Wentao Guo and Jikai Long and Yimeng Zeng and Zirui Liu and Xinyu Yang and Yide Ran and Jacob R. Gardner and Osbert Bastani and Christopher De Sa and Xiaodong Yu and Beidi Chen and Zhaozhuo Xu},
booktitle={The Thirteenth International Conference on Learning Representations},
year={2025}
}

@inproceedings{zhang2024revisitzo,
author = {Zhang, Yihua and Li, Pingzhi and Hong, Junyuan and Li, Jiaxiang and Zhang, Yimeng and Zheng, Wenqing and Chen, Pin-Yu and Lee, Jason D. and Yin, Wotao and Hong, Mingyi and Wang, Zhangyang and Liu, Sijia and Chen, Tianlong},
title = {Revisiting zeroth-order optimization for memory-efficient LLM fine-tuning: a benchmark},
year = {2024},
publisher = {JMLR.org},
booktitle = {Proceedings of the 41st International Conference on Machine Learning},
articleno = {2444},
numpages = {18},
location = {Vienna, Austria},
series = {ICML'24}
}

@article{clark2018arc,
  title={Think you have solved question answering? try arc, the ai2 reasoning challenge},
  author={Clark, Peter and Cowhey, Isaac and Etzioni, Oren and Khot, Tushar and Sabharwal, Ashish and Schoenick, Carissa and Tafjord, Oyvind},
  journal={arXiv preprint arXiv:1803.05457},
  year={2018}
}

@inproceedings{zellers2019hellaswag,
  title={Hellaswag: Can a machine really finish your sentence?},
  author={Zellers, Rowan and Holtzman, Ari and Bisk, Yonatan and Farhadi, Ali and Choi, Yejin},
  booktitle={Proceedings of the 57th annual meeting of the association for computational linguistics},
  pages={4791--4800},
  year={2019}
}

@article{sakaguchi2021winogrande,
  title={Winogrande: An adversarial winograd schema challenge at scale},
  author={Sakaguchi, Keisuke and Bras, Ronan Le and Bhagavatula, Chandra and Choi, Yejin},
  journal={Communications of the ACM},
  volume={64},
  number={9},
  pages={99--106},
  year={2021},
  publisher={ACM New York, NY, USA}
}

@inproceedings{mihaylov2018openbook,
  title={Can a suit of armor conduct electricity? a new dataset for open book question answering},
  author={Mihaylov, Todor and Clark, Peter and Khot, Tushar and Sabharwal, Ashish},
  booktitle={Proceedings of the 2018 conference on empirical methods in natural language processing},
  pages={2381--2391},
  year={2018}
}

@article{paszke2019pytorch,
  title={Pytorch: An imperative style, high-performance deep learning library},
  author={Paszke, Adam and Gross, Sam and Massa, Francisco and Lerer, Adam and Bradbury, James and Chanan, Gregory and Killeen, Trevor and Lin, Zeming and Gimelshein, Natalia and Antiga, Luca and others},
  journal={Advances in neural information processing systems},
  volume={32},
  year={2019}
}

@misc{zhao2024galore,
      title={GaLore: Memory-Efficient LLM Training by Gradient Low-Rank Projection}, 
      author={Jiawei Zhao and Zhenyu Zhang and Beidi Chen and Zhangyang Wang and Anima Anandkumar and Yuandong Tian},
      year={2024},
      eprint={2403.03507},
      archivePrefix={arXiv},
      primaryClass={cs.LG}
}

@inproceedings{
zhu2025apollo,
title={{APOLLO}: {SGD}-like Memory, AdamW-level Performance},
author={Hanqing Zhu and Zhenyu Zhang and Wenyan Cong and Xi Liu and Sem Park and Vikas Chandra and Bo Long and David Z. Pan and Zhangyang Wang and Jinwon Lee},
booktitle={Eighth Conference on Machine Learning and Systems},
year={2025}
}

@inproceedings{liu2024dora,
  title={Dora: Weight-decomposed low-rank adaptation},
  author={Liu, Shih-Yang and Wang, Chien-Yi and Yin, Hongxu and Molchanov, Pavlo and Wang, Yu-Chiang Frank and Cheng, Kwang-Ting and Chen, Min-Hung},
  booktitle={Forty-first International Conference on Machine Learning},
  year={2024}
}

@article{gautam2024variance,
  title={Variance-reduced zeroth-order methods for fine-tuning language models},
  author={Gautam, Tanmay and Park, Youngsuk and Zhou, Hao and Raman, Parameswaran and Ha, Wooseok},
  journal={arXiv preprint arXiv:2404.08080},
  year={2024}
}

@inproceedings{zhang2025mazo,
  title={Mazo: Masked zeroth-order optimization for multi-task fine-tuning of large language models},
  author={Zhang, Zhen and Yang, Yifan and Zhen, Kai and Susanj, Nathan and Mouchtaris, Athanasios and Kunzmann, Siegfried and Zhang, Zheng},
  booktitle={Proceedings of the 2025 Conference on Empirical Methods in Natural Language Processing},
  pages={18537--18554},
  year={2025}
}

@article{lin2026agzo,
  title={AGZO: Activation-Guided Zeroth-Order Optimization for LLM Fine-Tuning},
  author={Lin, Wei and Jiang, Yining and Song, Qingyu and Xiang, Qiao and Xu, Hong},
  journal={arXiv preprint arXiv:2601.17261},
  year={2026}
}

\clearpage
\appendix
\onecolumn
\newpage
\appendix
\onecolumn

% \startcontents[appendix]

% \phantomsection
% \section*{\centering Appendix Contents}
% \addcontentsline{toc}{section}{Appendix Contents}

% \vspace{0.5em}

% % Automatically prints all subsequent appendix sections and subsections.
% \printcontents[appendix]{}{1}{%
% \setcounter{tocdepth}{3}%
% }

\clearpage

\section{Proofs}

\subsection{Notation and Gradient Energy}

Table~\ref{tab:notations} summarizes the notation. The support-selection analysis considers one linear layer. The alignment and convergence results use the total selected dimension across the optimized parameter blocks.

\begin{table}[ht]
\centering
\caption{Summary of notation.}
\label{tab:notations}
\renewcommand{\arraystretch}{1.25}
\begin{tabular}{ll}
\toprule
\textbf{Symbol} & \textbf{Description} \\
\midrule
$\mathbf{W},\mathbf{M},\mathbf{Z}$ & Weights, binary support mask, and Gaussian perturbation. \\
$\mathbf{G},\hat{\mathbf{G}}_\epsilon,\hat{\mathbf{G}}_0$ & Stochastic gradient, finite-difference estimate, and its directional limit. \\
$\mathbf{x},\boldsymbol{\delta}$ & Per-observation input and backpropagated signal. \\
$\mathbf{X}\in\mathbb{R}^{n\times B}$ & Matrix of calibration activation observations for one layer. \\
$v_j,\hat v_j$ & Population and empirical second moments of feature $j$. \\
$\mathbf{S}_{i,j}$ & Wanda score of weight $\mathbf{W}_{i,j}$. \\
$\mathcal{I}_{\mathrm{WIS},i},\mathcal{I}_i^*,\mathcal{A}_i$ & WIS support, oracle support, and their union in row $i$. \\
$r,\;k=rn$ & Perturbation density and selected coordinates per row of a layer. \\
$d=\sum_\ell m_\ell n_\ell$ & Total number of parameters in the optimized blocks. \\
$s=d_{\mathrm{eff}}=\sum_\ell m_\ell k_\ell$ & Total number of selected perturbation coordinates. \\
$\gamma,\nu$ & Bounds on input--error dependence. \\
$K,\xi$ & Normalized sub-Gaussian constant and calibration error tolerance. \\
$\kappa_i,\kappa_{i,\xi},\kappa_\xi$ & Row-wise comparison factors and their minimum. \\
$\rho,\rho^*,\rho_{\min}$ & Captured energy, oracle capture, and a trajectory-wise lower bound. \\
$C_{i,\mathrm{low}},C_{i,\mathrm{high}}$ & Gradient--saliency comparison constants. \\
$\epsilon,\eta,\tau$ & Perturbation radius, learning rate, and mask refresh interval. \\
$L,\mu,\sigma^2$ & Smoothness, PL, and stochastic-variance constants. \\
$a=s+2,\;\sigma_\epsilon^2$ & Gaussian moment factor and variance bound including finite differences. \\
\midrule
$\|\cdot\|_2,\|\cdot\|_F$ & Euclidean and Frobenius norms. \\
$\langle\cdot,\cdot\rangle,\odot$ & Frobenius inner product and element-wise multiplication. \\
\bottomrule
\end{tabular}
\end{table}

For a linear layer $\mathbf{y}=\mathbf{W}\mathbf{x}$, the per-example gradient satisfies
\begin{equation}
\mathbf{G}_{i,j}
=\frac{\partial f}{\partial\mathbf{W}_{i,j}}
=\delta_i x_j,
\qquad
\delta_i=\frac{\partial f}{\partial y_i}.
\nonumber
\end{equation}
The calibration analysis concerns $\mathbb{E}[\mathbf{G}_{i,j}^2]$ over the downstream distribution at a fixed parameter value. For $B$ calibration observations,
\begin{equation}
\hat v_j=\frac1B\sum_{b=1}^B(x_j^{(b)})^2,
\qquad
v_j=\mathbb{E}[x_j^2],
\qquad
\mathbf{S}_{i,j}^2=B\mathbf{W}_{i,j}^2\hat v_j.
\nonumber
\end{equation}
The following proofs establish the energy-capture and convergence results in the main text.

\subsection{Proof of Lemma~\ref{lemma:gradient_saliency}}

\begin{proof}
Assumption~\ref{assump:weak_data_corre} gives
\begin{equation}
\gamma\mathbb{E}[\delta_i^2]v_j
\leq\mathbb{E}[\mathbf{G}_{i,j}^2]
\leq\nu\mathbb{E}[\delta_i^2]v_j.
\label{eq:grad_decoupled}
\end{equation}
Under Assumption~\ref{assump:sub_gaussian_act}, the squared activation is sub-exponential with scale bounded by a constant multiple of $Kv_j$. For features with $v_j>0$, concentration of the empirical mean gives, for a universal constant $c>0$ and $\xi\in(0,1)$,
\begin{equation}
\Pr(|\hat v_j-v_j|\geq\xi v_j)
\leq2\exp\!\left(-\frac{cB\xi^2}{K^2}\right).
\nonumber
\end{equation}
Features with $v_j=0$ satisfy $\hat v_j=0$ almost surely. A union bound over the $n$ features yields
\begin{equation}
(1-\xi)v_j\leq\hat v_j\leq(1+\xi)v_j,
\qquad j=1,\ldots,n,
\nonumber
\end{equation}
with probability at least $1-\mathcal{O}(n^{-1})$ when $B=\Omega(K^2\log n/\xi^2)$. On this event,
\begin{equation}
\frac{\mathbf{S}_{i,j}^2}{B(1+\xi)\mathbf{W}_{i,j}^2}
\leq v_j
\leq\frac{\mathbf{S}_{i,j}^2}{B(1-\xi)\mathbf{W}_{i,j}^2}.
\label{eq:vj_concentration}
\end{equation}
For every nonzero-energy row and $j\in\mathcal{A}_i$, combining Eqs.~\eqref{eq:grad_decoupled} and~\eqref{eq:vj_concentration} gives
\begin{equation}
\frac{\gamma\mathbb{E}[\delta_i^2]}{B(1+\xi)w_{\max,i}}\mathbf{S}_{i,j}^2
\leq\mathbb{E}[\mathbf{G}_{i,j}^2]
\leq\frac{\nu\mathbb{E}[\delta_i^2]}{B(1-\xi)w_{\min,i}}\mathbf{S}_{i,j}^2.
\nonumber
\end{equation}
The two coefficients define $C_{i,\mathrm{low}}$ and $C_{i,\mathrm{high}}$, and
\begin{equation}
\frac{C_{i,\mathrm{low}}}{C_{i,\mathrm{high}}}
=\frac{\gamma w_{\min,i}}{\nu w_{\max,i}}\frac{1-\xi}{1+\xi}
=\kappa_{i,\xi}.
\nonumber
\end{equation}
Rows with zero gradient energy contribute zero to the global ratio.
\end{proof}

\subsection{Proof of Theorem~\ref{theo:gradient_energy}}

\begin{proof}
Define the row-wise oracle over the full row by
\begin{equation}
\mathcal{I}_i^*\in
\operatorname*{arg\,max}_{\mathcal{I}\subseteq\{1,\ldots,n\},\;|\mathcal{I}|=k}
\sum_{j\in\mathcal{I}}\mathbb{E}[\mathbf{G}_{i,j}^2].
\nonumber
\end{equation}
The WIS support maximizes the sum of squared saliency scores among all sets of size $k$. On the concentration event in Lemma~\ref{lemma:gradient_saliency}, both supports belong to $\mathcal{A}_i$, so
\begin{equation}
\begin{aligned}
\sum_{j\in\mathcal{I}_{\mathrm{WIS},i}}\mathbb{E}[\mathbf{G}_{i,j}^2]
&\geq C_{i,\mathrm{low}}\sum_{j\in\mathcal{I}_{\mathrm{WIS},i}}\mathbf{S}_{i,j}^2
\\
&\geq C_{i,\mathrm{low}}\sum_{j\in\mathcal{I}_i^*}\mathbf{S}_{i,j}^2
\\
&\geq\kappa_{i,\xi}\sum_{j\in\mathcal{I}_i^*}\mathbb{E}[\mathbf{G}_{i,j}^2].
\end{aligned}
\nonumber
\end{equation}
Dividing by the total expected energy in the row and summing with weights $\omega_i$ gives
\begin{equation}
\rho_{\mathrm{WIS}}
=\sum_i\omega_i\rho_{\mathrm{WIS},i}
\geq\sum_i\omega_i\kappa_{i,\xi}\rho_i^*
\geq\kappa_\xi\rho^*.
\nonumber
\end{equation}
For a uniform row-wise mask independent of the gradient, each coordinate is included with probability $r$. Therefore,
\begin{equation}
\mathbb{E}_{\mathbf{M}_{\mathrm{rand}}}[\rho(\mathbf{M}_{\mathrm{rand}})]
=\frac{\sum_{i,j}r\mathbb{E}[\mathbf{G}_{i,j}^2]}{\mathbb{E}[\|\mathbf{G}\|_F^2]}
=r.
\nonumber
\end{equation}
The sufficient condition $\kappa_\xi\rho^*>r$ then gives the stated comparison.
\end{proof}

\subsection{Proof of Lemma~\ref{lemma:cosine_sim}}

\begin{proof}
Fix the parameter value and mask. Write $\mathbf{U}=(\mathbf{Z}\odot\mathbf{M})/\sqrt r$, where the entries of $\mathbf{Z}$ are independent standard Gaussians, and $\hat{\mathbf{G}}_0=\langle\mathbf{G},\mathbf{U}\rangle\mathbf{U}$. Taking the perturbation expectation first gives
\begin{equation}
\begin{aligned}
\mathbb{E}[\langle\hat{\mathbf{G}}_0,\mathbf{G}\rangle]
&=\mathbb{E}_{\mathcal{B}}[\langle\mathbb{E}_{\mathbf{Z}}[\hat{\mathbf{G}}_0\mid\mathbf{G}],\mathbf{G}\rangle]
\\
&=\frac1r\mathbb{E}_{\mathcal{B}}[\|\mathbf{M}\odot\mathbf{G}\|_F^2]
\\
&=\frac\rho r\mathbb{E}_{\mathcal{B}}[\|\mathbf{G}\|_F^2].
\end{aligned}
\label{eq:inner_prod_exact}
\end{equation}
The Gaussian fourth-moment identity in the $s$ selected coordinates yields
\begin{equation}
\mathbb{E}_{\mathbf{Z}}[\|\hat{\mathbf{G}}_0\|_F^2\mid\mathbf{G}]
=\frac{s+2}{r^2}\|\mathbf{M}\odot\mathbf{G}\|_F^2.
\nonumber
\end{equation}
Hence,
\begin{equation}
\mathbb{E}[\|\hat{\mathbf{G}}_0\|_F^2]
=\frac{s+2}{r^2}\rho\mathbb{E}[\|\mathbf{G}\|_F^2].
\label{eq:norm_est_bound}
\end{equation}
Substituting Eqs.~\eqref{eq:inner_prod_exact} and~\eqref{eq:norm_est_bound} into the moment-normalized alignment gives
\begin{equation}
\operatorname{Align}(\hat{\mathbf{G}}_0,\mathbf{G})
=\sqrt{\frac\rho{s+2}}.
\nonumber
\end{equation}
For a uniform random mask independent of this gradient distribution, taking an additional expectation over masks replaces $\rho$ by $r$. The corresponding ratio of moment-normalized alignment values is $\sqrt{\rho/r}$.
\end{proof}

\subsection{Proof of Theorem~\ref{theo:convergence}}

\begin{proof}
Condition on the fixed calibration mask and the optimization history before iteration $t$. Write $\mathbb{E}_t$ for expectation over the current mini-batch and perturbation, and define
\begin{equation}
\mathbf{H}_t=\mathbf{M}\odot\nabla f(\mathbf{W}^t),
\qquad
\mathbf{U}=\frac{\mathbf{Z}\odot\mathbf{M}}{\sqrt r},
\qquad
a=s+2.
\nonumber
\end{equation}
The actual central-difference estimator and its directional limit are
\begin{equation}
\hat{\mathbf{G}}_\epsilon^t
=\frac{f(\mathbf{W}^t+\epsilon\mathbf{U},\mathcal{B}^t)-f(\mathbf{W}^t-\epsilon\mathbf{U},\mathcal{B}^t)}{2\epsilon}\mathbf{U},
\qquad
\hat{\mathbf{G}}_0^t=\langle\mathbf{G}^t,\mathbf{U}\rangle\mathbf{U}.
\nonumber
\end{equation}
Gaussian symmetry and integration by parts give
\begin{equation}
\mathbb{E}_t[\hat{\mathbf{G}}_\epsilon^t]
=\frac1r\mathbf{M}\odot\mathbb{E}_{\mathbf{Z}}[\nabla f(\mathbf{W}^t+\epsilon\mathbf{U})].
\nonumber
\end{equation}
For $\mathbf{b}_t=\mathbb{E}_t[\hat{\mathbf{G}}_\epsilon^t]-\mathbf{H}_t/r$, smoothness and $\mathbb{E}\|\mathbf{U}\|_F^2=s/r$ imply
\begin{equation}
\begin{aligned}
\|\mathbf{b}_t\|_F^2
&\leq\frac1{r^2}\mathbb{E}_{\mathbf{Z}}[\|\mathbf{M}\odot(\nabla f(\mathbf{W}^t+\epsilon\mathbf{U})-\nabla f(\mathbf{W}^t))\|_F^2]
\\
&\leq\frac{L^2\epsilon^2}{r^2}\mathbb{E}_{\mathbf{Z}}[\|\mathbf{U}\|_F^2]
\\
&=\frac{L^2\epsilon^2s}{r^3}.
\end{aligned}
\label{eq:finite_difference_bias}
\end{equation}
For each mini-batch, the fundamental theorem of calculus and Assumption~\ref{assump:L_lipschitz} give
\begin{equation}
\begin{aligned}
\|\hat{\mathbf{G}}_\epsilon^t-\hat{\mathbf{G}}_0^t\|_F
&\leq\frac{\|\mathbf{U}\|_F}{2\epsilon}\int_{-\epsilon}^{\epsilon}
\left|\left\langle\nabla f(\mathbf{W}^t+v\mathbf{U},\mathcal{B}^t)-\mathbf{G}^t,\mathbf{U}\right\rangle\right|\,dv
\\
&\leq\frac{L\|\mathbf{U}\|_F^3}{2\epsilon}\int_{-\epsilon}^{\epsilon}|v|\,dv
\\
&=\frac{L\epsilon}{2}\|\mathbf{U}\|_F^3.
\end{aligned}
\nonumber
\end{equation}
Since $r\|\mathbf{U}\|_F^2$ has a chi-squared distribution with $s$ degrees of freedom,
\begin{equation}
\mathbb{E}_t[\|\hat{\mathbf{G}}_\epsilon^t-\hat{\mathbf{G}}_0^t\|_F^2]
\leq\frac{L^2\epsilon^2}{4}\mathbb{E}[\|\mathbf{U}\|_F^6]
=\frac{L^2\epsilon^2s(s+2)(s+4)}{4r^3}.
\label{eq:finite_difference_remainder}
\end{equation}
By Lemma~\ref{lemma:cosine_sim} and the conditional variance bound,
\begin{equation}
\begin{aligned}
\mathbb{E}_t[\|\hat{\mathbf{G}}_0^t\|_F^2]
&=\frac a{r^2}\mathbb{E}_t[\|\mathbf{M}\odot\mathbf{G}^t\|_F^2]
\\
&\leq\frac a{r^2}(\|\mathbf{H}_t\|_F^2+\sigma^2).
\end{aligned}
\nonumber
\end{equation}
Combining this inequality with Eq.~\eqref{eq:finite_difference_remainder} and $\|\mathbf{A}+\mathbf{B}\|_F^2\leq2\|\mathbf{A}\|_F^2+2\|\mathbf{B}\|_F^2$ yields
\begin{equation}
\mathbb{E}_t[\|\hat{\mathbf{G}}_\epsilon^t\|_F^2]
\leq\frac{2a}{r^2}(\|\mathbf{H}_t\|_F^2+\sigma_\epsilon^2),
\qquad
\sigma_\epsilon^2=\sigma^2+\frac{L^2\epsilon^2s(s+4)}{4r}.
\label{eq:noise_term_new}
\end{equation}
Both the estimator and its bias are supported on $\mathbf{M}$. Applying Young's inequality gives
\begin{equation}
\begin{aligned}
\mathbb{E}_t[\langle\nabla f(\mathbf{W}^t),\hat{\mathbf{G}}_\epsilon^t\rangle]
&=\frac1r\|\mathbf{H}_t\|_F^2+\langle\mathbf{H}_t,\mathbf{b}_t\rangle
\\
&\geq\frac3{4r}\|\mathbf{H}_t\|_F^2-r\|\mathbf{b}_t\|_F^2.
\end{aligned}
\label{eq:signal_term_new}
\end{equation}
Using $\mathbf{W}^{t+1}=\mathbf{W}^t-\eta_t\hat{\mathbf{G}}_\epsilon^t$ and population smoothness,
\begin{equation}
\mathbb{E}_t[f(\mathbf{W}^{t+1})]
\leq f(\mathbf{W}^t)
-\eta_t\mathbb{E}_t[\langle\nabla f(\mathbf{W}^t),\hat{\mathbf{G}}_\epsilon^t\rangle]
+\frac{L\eta_t^2}{2}\mathbb{E}_t[\|\hat{\mathbf{G}}_\epsilon^t\|_F^2].
\nonumber
\end{equation}
For $\eta_t\leq r/(4La)$, substituting Eqs.~\eqref{eq:finite_difference_bias}, \eqref{eq:noise_term_new}, and~\eqref{eq:signal_term_new} gives
\begin{equation}
\mathbb{E}_t[f(\mathbf{W}^{t+1})]
\leq f(\mathbf{W}^t)
-\frac{\eta_t}{2r}\|\mathbf{H}_t\|_F^2
+\frac{La\eta_t^2}{r^2}\sigma_\epsilon^2
+\frac{L^2\epsilon^2s\eta_t}{r^2}.
\label{eq:finite_difference_descent}
\end{equation}
Now set $\eta_t=r/(4La\sqrt{t+1})$. Taking full expectations, summing, and using $f\geq f^*$ and the deterministic lower bound on captured energy yield
\begin{equation}
\begin{aligned}
\rho_{\min}\sum_{t=0}^{T-1}\frac{\mathbb{E}[\|\nabla f(\mathbf{W}^t)\|_F^2]}{\sqrt{t+1}}
&\leq8La\Delta f
+\frac{\sigma_\epsilon^2}{2}\sum_{t=0}^{T-1}\frac1{t+1}
+\frac{2L^2\epsilon^2s}{r}\sum_{t=0}^{T-1}\frac1{\sqrt{t+1}}
\\
&\leq8La\Delta f+\frac{\sigma_\epsilon^2}{2}(1+\log T)
+\frac{4L^2\epsilon^2s\sqrt T}{r}.
\end{aligned}
\nonumber
\end{equation}
Finally, $1/\sqrt{t+1}\geq1/\sqrt T$ for $t<T$, so
\begin{equation}
\frac1T\sum_{t=0}^{T-1}\mathbb{E}[\|\nabla f(\mathbf{W}^t)\|_F^2]
\leq\frac{8La\Delta f+\tfrac12\sigma_\epsilon^2(1+\log T)}{\rho_{\min}\sqrt T}
+\frac{4L^2\epsilon^2s}{r\rho_{\min}}.
\nonumber
\end{equation}
\end{proof}

\subsection{Convergence under the PL Condition}

The same descent inequality gives geometric convergence to a noise and finite-difference neighborhood under a PL condition.

\begin{assumption}[$\mu$-PL Condition]\label{assump:pl_condition}
For $0<\mu\leq L$,
\begin{equation}
\|\nabla f(\mathbf{W})\|_F^2\geq2\mu(f(\mathbf{W})-f^*).
\nonumber
\end{equation}
\end{assumption}

\begin{theorem}[Convergence under the PL Condition]\label{theo:pl_convergence}
Suppose Assumptions~\ref{assump:L_lipschitz}, \ref{assump:bounded_variance}, and~\ref{assump:pl_condition} hold, together with the fixed-support energy condition in Theorem~\ref{theo:convergence}. Let $a=s+2$, retain $\sigma_\epsilon^2$ from that theorem, and take $\eta=\alpha r/(4La)$ with $\alpha\in(0,1]$. Then
\begin{equation}
\mathbb{E}[f(\mathbf{W}^T)-f^*]
\leq\left(1-\frac{\alpha\mu\rho_{\min}}{4La}\right)^T\Delta f
+\frac{\alpha\sigma_\epsilon^2}{4\mu\rho_{\min}}
+\frac{L^2\epsilon^2s}{r\mu\rho_{\min}}.
\nonumber
\end{equation}
\end{theorem}

\begin{proof}
Write $\Delta_t=\mathbb{E}[f(\mathbf{W}^t)-f^*]$. Substituting the constant step size into Eq.~\eqref{eq:finite_difference_descent} and applying the support-energy and PL conditions gives
\begin{equation}
\Delta_{t+1}
\leq\left(1-\frac{\alpha\mu\rho_{\min}}{4La}\right)\Delta_t
+\frac{\alpha^2\sigma_\epsilon^2}{16La}
+\frac{\alpha L\epsilon^2s}{4ar}.
\nonumber
\end{equation}
Set $q=1-\alpha\mu\rho_{\min}/(4La)\in[0,1)$. Iterating this inequality yields
\begin{equation}
\begin{aligned}
\Delta_T
&\leq q^T\Delta_0
+\left(\frac{\alpha^2\sigma_\epsilon^2}{16La}+\frac{\alpha L\epsilon^2s}{4ar}\right)\sum_{t=0}^{T-1}q^t
\\
&\leq q^T\Delta_0
+\frac{\alpha\sigma_\epsilon^2}{4\mu\rho_{\min}}
+\frac{L^2\epsilon^2s}{r\mu\rho_{\min}}.
\end{aligned}
\nonumber
\end{equation}
\end{proof}

\section{Experimental Details}
\label{app:experimental_details}

For the GLUE, SQuAD, and SuperGLUE experiments, all ZO methods train for 20,000 steps with a batch size of 16; FT and LoRA use a batch size of 8. Checkpoints are selected by validation performance. We use a sequence length of 2048, constant learning rates, and zero weight decay for ZO methods, following the settings in MeZO and Sparse-MeZO~\cite{malladi2023_mezo,liu2025sparse}.

Tables~\ref{tab:tested_hyperparameters}--\ref{tab:super_glue_hyper} report the hyperparameter search and selected configurations. Results on GLUE and SQuAD are averaged over five seeds, while SuperGLUE results are from one run. WIS-MeZO uses 128 calibration examples from downstream training data and a fixed mask.

\begin{table}[!th]
    \centering
    \caption{The hyperparameter grids used for the experiments in the main text.}
    \label{tab:tested_hyperparameters}
    \begin{tabular}{c|cc}
    \toprule
       Methods  &  Hyperparameters & Values \\
       \midrule
         \multirow{3}{*}{FT} & Batch size & 8 \\
         & Learning rate & $\{1.0e-6, 5.0e-6,1.0e-5,5.0e-5\}$\\
         & $(\beta_1,\beta_2)$ & (0.9, 0.999) \\
         \midrule
         \multirow{5}{*}{LoRA} & Batch size & 8 \\
         & Learning rate & $\{1.0e-5, 5.0e-5,1.0e-4,5.0e-4\}$\\
         & Rank & 8 \\
         & Alpha-$\alpha$ & 16 \\
         & $(\beta_1,\beta_2)$ & (0.9, 0.999) \\
         \midrule 
         \multirow{3}{*}{MeZO} &  Batch size & 16 \\
         & Learning rate & $\{5.0e-8,1.0e-7,2.5e-7,5.0e-7\}$\\
         & $\epsilon$ & $\{1.0e-3,1.0e-4\}$\\
         \midrule
         \multirow{5}{*}{LOZO} &  Batch size & 16 \\
         & Learning rate & $\{5.0e-8,1.0e-7,2.5e-7,5.0e-7\}$\\
         & $\epsilon$ & $\{1.0e-3,1.0e-4\}$\\
         & Rank & \{1, 2\} \\
         & Interval & 50 \\
         \midrule 
         \multirow{4}{*}{Sparse-MeZO} &  Batch size & 16 \\
         & Learning rate & $\{5.0e-7,1.0e-6,2.5e-6,5.0e-6\}$ \\
         & $\epsilon$ & $\{1.0e-3,1.0e-4\}$\\
         & Density ratio-$r$ & $\{0.05, 0.1, 0.2, 0.4\}$\\
         \midrule 
         \multirow{4}{*}{Random-MeZO} &  Batch size & 16 \\
         & Learning rate & $\{5.0e-8,1.0e-7,2.5e-7,5.0e-7\}$ \\
         & $\epsilon$ & $\{1.0e-3,1.0e-4\}$\\
         & Density ratio-$r$ & $\{0.05, 0.1, 0.2, 0.4\}$\\
         \midrule 
         \multirow{4}{*}{WIS-MeZO} &  Batch size & 16 \\
         & Learning rate & $\{5.0e-8,1.0e-7,2.5e-7,5.0e-7\}$ \\
         & $\epsilon$ & $\{1.0e-3,1.0e-4\}$\\
         & Density ratio-$r$ & $\{0.05, 0.1, 0.2, 0.4\}$\\
         \midrule
         \multirow{5}{*}{WISD-MeZO} &  Batch size & 16 \\
         & Learning rate & $\{5.0e-8,1.0e-7,2.5e-7,5.0e-7\}$ \\
         & $\epsilon$ & $\{1.0e-3,1.0e-4\}$\\
         & Density ratio-$r$ & $\{0.05, 0.1, 0.2, 0.4\}$\\
         & Frequency-$\tau$ & $\{2000,5000,10000\}$\\
         \bottomrule
    \end{tabular}
\end{table}

\begin{table*}[!th]
    \centering
    \caption{Selected hyperparameters for the GLUE and SQuAD experiments.}
    \label{tab:glu_qwen_llama_hyper}
    \resizebox{\linewidth}{!}{\begin{tabular}{l|l|cccccc}
    \toprule
       Models & Methods & \textbf{MNLI} & \textbf{RTE} & \textbf{QQP} & \textbf{MRPC} & \textbf{SST-2} & \textbf{SQuAD} \\
       \midrule
       \multirow{8}{*}{LLaMA3.2-3B} & FT & 1.0e-5 & 1.0e-5 & 1.0e-5 & 1.0e-5 & 1.0e-5 & 1.0e-5 \\
       & LoRA & 5.0e-4 & 5.0e-4 & 5.0e-4 & 5.0e-4 & 1.0e-4 & 5.0e-5 \\
       % \midrule
       & MeZO & 5.0e-7 &  2.5e-7 & 5.0e-7 & 2.5e-7 & 5.0e-7 & 2.5e-7 \\
       & LOZO & 1.0e-7 &  2.5e-7 & 1.0e-7 & 2.5e-7 & 2.5e-7 & 2.5e-7 \\
       & Sparse-MeZO & 2.5e-6, 0.2 & 1.0e-6, 0.2 & 1.0e-6, 0.4 & 1.0e-6, 0.4 & 1.0e-6, 0.4 & 1.0e-6, 0.2 \\
       & Random-MeZO & 5.0e-7, 0.2 & 5.0e-7, 0.2 & 5.0e-7, 0.4 & 5.0e-7, 0.2 & 5.0e-7, 0.2 & 5.0e-7, 0.2 \\
       \cmidrule(){2-8}
       & WIS-MeZO & 2.5e-7, 0.2 & 2.5e-7, 0.2 & 5.0e-7, 0.4 & 2.5e-7, 0.1 & 1.0e-7, 0.1 & 1.0e-7, 0.1 \\
       & WISD-MeZO & 2.5e-7, 0.2, 5000 & 2.5e-7, 0.1, 2000 & 2.5e-7, 0.2, 10000& 2.5e-7, 0.1, 2000 & 1.0e-7, 0.1, 5000 & 1.0e-7, 0.1, 5000 \\
       \midrule 
       \multirow{8}{*}{Qwen2.5-3B} & FT & 1.0e-5 & 1.0e-5 & 1.0e-5 & 1.0e-5 & 1.0e-5 & 1.0e-5 \\
       & LoRA & 1.0e-4 & 1.0e-4 & 5.0e-5 & 1.0e-4 & 1.0e-5 & 5.0e-4 \\
       & MeZO & 1.0e-7 & 1.0e-7 & 2.5e-7 & 5.0e-7 & 5.0e-7 & 5.0e-7 \\
       & LOZO & 5.0e-8 & 1.0e-7 & 1.0e-7 & 1.0e-7 & 2.5e-7 & 2.5e-7 \\
       & Sparse-MeZO & 2.5e-6, 0.1 & 2.5e-6, 0.4 & 5.0e-6, 0.1 & 5.0e-6, 0.2 & 5.0e-6, 0.4 & 5.0e-6, 0.1 \\
       & Random-MeZO & 1.0e-7, 0.1 & 5.0e-7, 0.4 & 1.0e-7, 0.1 & 2.5e-7, 0.2 & 5.0e-7, 0.4 & 5.0e-7, 0.4 \\
       \cmidrule(){2-8}
       & WIS-MeZO & 1.0e-7, 0.2 & 1.0e-7, 0.1 & 5.0e-8, 0.4 & 5.0e-8, 0.2 & 2.5e-7, 0.4 & 5.0e-8, 0.05 \\
       & WISD-MeZO & 5.0e-8, 0.2, 5000 & 5.0e-8, 0.05, 10000 & 5.0e-8, 0.4, 2000 & 5.0e-8, 0.2, 5000 & 2.5e-7, 0.2, 2000 & 5.0e-8, 0.05, 5000 \\
       \bottomrule
    \end{tabular}}
\end{table*}

\begin{table*}[!th]
    \centering
    \caption{Selected hyperparameters for the SuperGLUE experiments.}
    \label{tab:super_glue_hyper}
    \resizebox{\linewidth}{!}{\begin{tabular}{l|l|ccccc}
    \toprule
       Models & Methods & \textbf{CB} & \textbf{MultiRC} & \textbf{WSC} & \textbf{ReCoRD} & \textbf{BoolQ} \\
       \midrule
       \multirow{8}{*}{LLaMA3.2-3B} & FT & 1.0e-5 & 1.0e-5 & 1.0e-5 & 1.0e-6 & 1.0e-5 \\
       & LoRA & 5.0e-4 & 1.0e-4 & 5.0e-4 & 1.0e-5 & 5.0e-4 \\
       & MeZO & 2.5e-7 & 5.0e-7 & 2.5e-7 & 2.5e-7 & 1.0e-7 \\
       & LOZO & 1.0e-7 & 2.5e-7 & 1.0e-7 & 2.5e-7 & 1.0e-7 \\
       & Sparse-MeZO & 1.0e-6, 0.4 & 1.0e-6, 0.2 & 5.0e-6, 0.4 & 1.0e-6, 0.2 & 2.5e-6, 0.2 \\
       & Random-MeZO & 5.0e-7, 0.2 & 5.0e-7, 0.2 & 2.5e-7, 0.1 & 5.0e-7, 0.4 & 5.0e-7, 0.2 \\
       \cmidrule(){2-7}
       & WIS-MeZO & 2.5e-7, 0.1 & 2.5e-7, 0.1 & 5.0e-7, 0.05 & 1.0e-7, 0.2 & 1.0e-7, 0.05 \\
       & WISD-MeZO & 2.5e-7,  0.1, 2000 & 2.5e-7, 0.1,5000 & 1.0e-7, 0.05,2000 & 1.0e-7, 0.2, 10000 & 1.0e-7, 0.1, 10000 \\
       \midrule 
       \multirow{8}{*}{Qwen2.5-3B} & FT & 5.0e-5 & 1.0e-5 & 1.0e-5 & 1.0e-6 & 1.0e-5 \\
       & LoRA & 5.0e-5 & 5.0e-5 & 1.0e-5 & 1.0e-5 & 1.0e-4 \\
       & MeZO & 5.0e-7 & 2.5e-7 & 2.5e-7 & 5.0e-7 & 2.5e-7 \\
       & LOZO & 1.0e-7 & 1.0e-7 & 1.0e-7 & 2.5e-7 & 1.0e-7 \\
       & Sparse-MeZO & 2.5e-6, 0.2 & 1.0e-6, 0.4 & 2.5e-6, 0.2 & 5.0e-6, 0.2 & 2.5e-6, 0.2 \\
       & Random-MeZO & 1.0e-7, 0.1 & 5.0e-7, 0.4 & 2.5e-7, 0.2 & 5.0e-7, 0.2 & 1.0e-7, 0.1 \\
       \cmidrule(){2-7}
       & WIS-MeZO & 1.0e-7, 0.1 & 5.0e-8, 0.05 & 1.0e-7, 0.2 & 1.0e-7, 0.1 & 1.0e-7, 0.2 \\
       & WISD-MeZO & 2.5e-7, 0.2, 2000 & 2.5e-7, 0.4, 5000 & 1.0e-7, 0.2, 10000 & 1.0e-7, 0.1, 2000 & 1.0e-7, 0.2,10000 \\
       \bottomrule
    \end{tabular}}
\end{table*}

Regarding the computing environment, we use Python 3.10.14 and PyTorch 2.8.0~\cite{paszke2019pytorch}. All experiments were conducted on NVIDIA H20 GPUs.

\section{Additional Gradient Energy Capture Results}
\label{app:additional_energy_capture}

We compare Random, W-small, W-large, and WIS supports at matched density. The supports are selected uniformly at random, by small weight magnitude, by large weight magnitude, and by the Wanda score, respectively. For each task, we evaluate their gradient energy capture on 2,000 gradient snapshots from an auxiliary AdamW training run on LLaMA-3.2-3B. The supports are fixed at initialization, and all strategies are evaluated on the same gradients.

For a support mask $\mathbf{M}$ and a mini-batch gradient $\mathbf{G}^t$, energy capture is measured by
\begin{equation}
\rho_t(\mathbf{M})
=\frac{\|\mathbf{M}\odot\mathbf{G}^t\|_F^2}{\|\mathbf{G}^t\|_F^2}.
\label{eq:empirical_energy_capture}
\end{equation}
We report the average over optimization steps. Random masking captures an expected fraction $r$ of gradient energy. At $r=0.05$, WIS preserves over $40\%$ of the energy and exceeds the other selection rules. W-small also outperforms W-large and random masking.

\begin{figure*}[!th]
    \centering

    \begin{subfigure}[b]{0.32\linewidth}
        \centering
        \includegraphics[width=\linewidth]{
            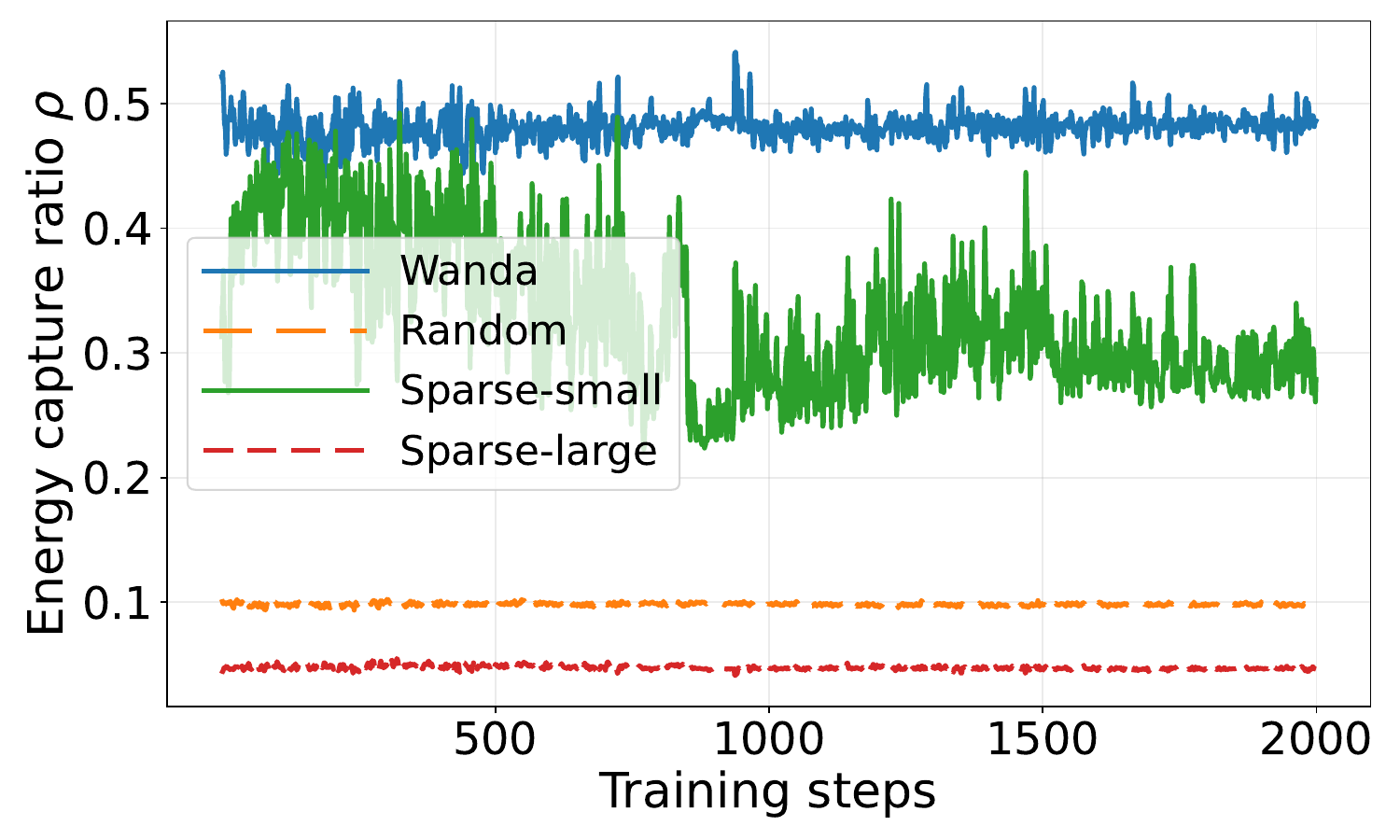
        }
        \caption{RTE}
    \end{subfigure}
    \hfill
    \begin{subfigure}[b]{0.32\linewidth}
        \centering
        \includegraphics[width=\linewidth]{
            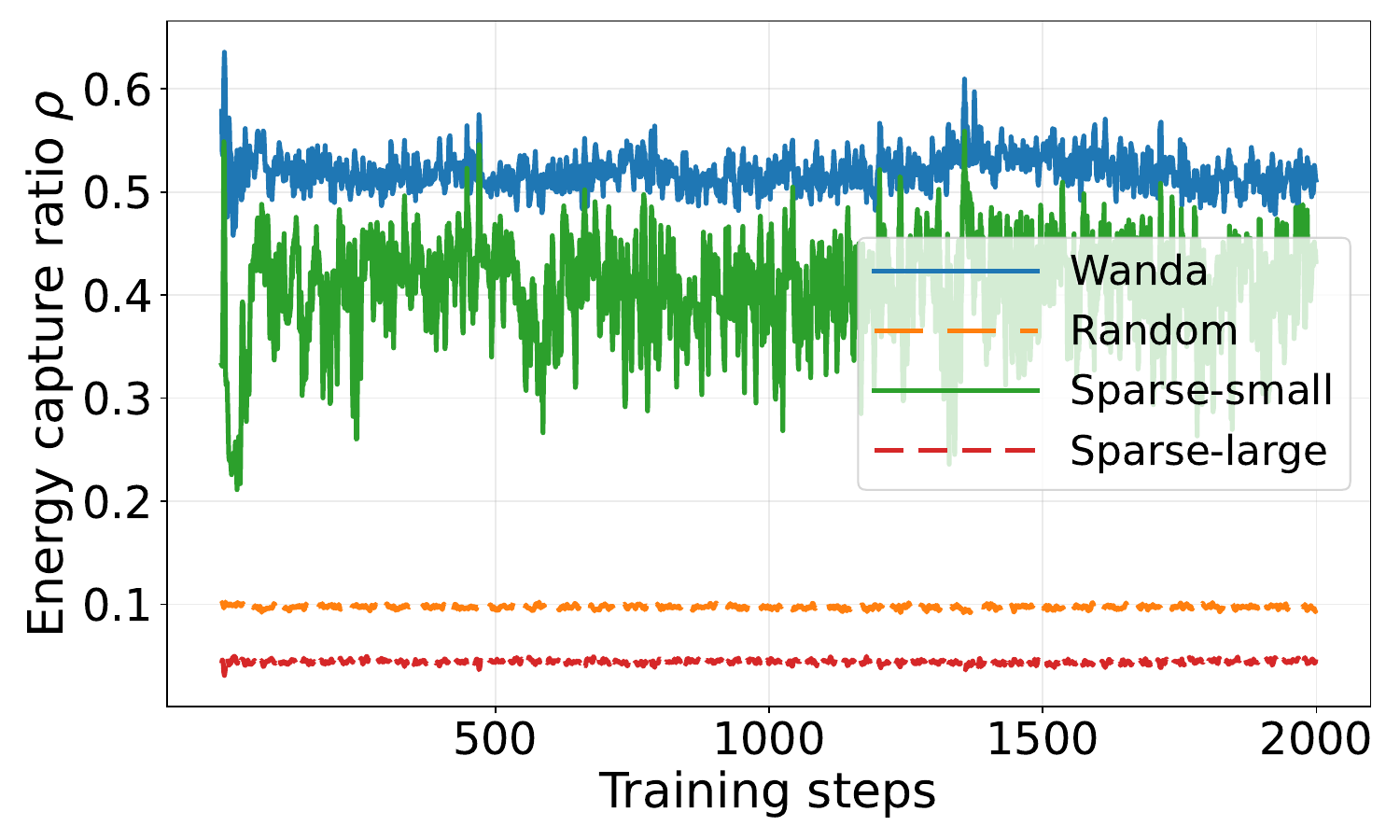
        }
        \caption{QQP}
    \end{subfigure}
    \hfill
    \begin{subfigure}[b]{0.32\linewidth}
        \centering
        \includegraphics[width=\linewidth]{
            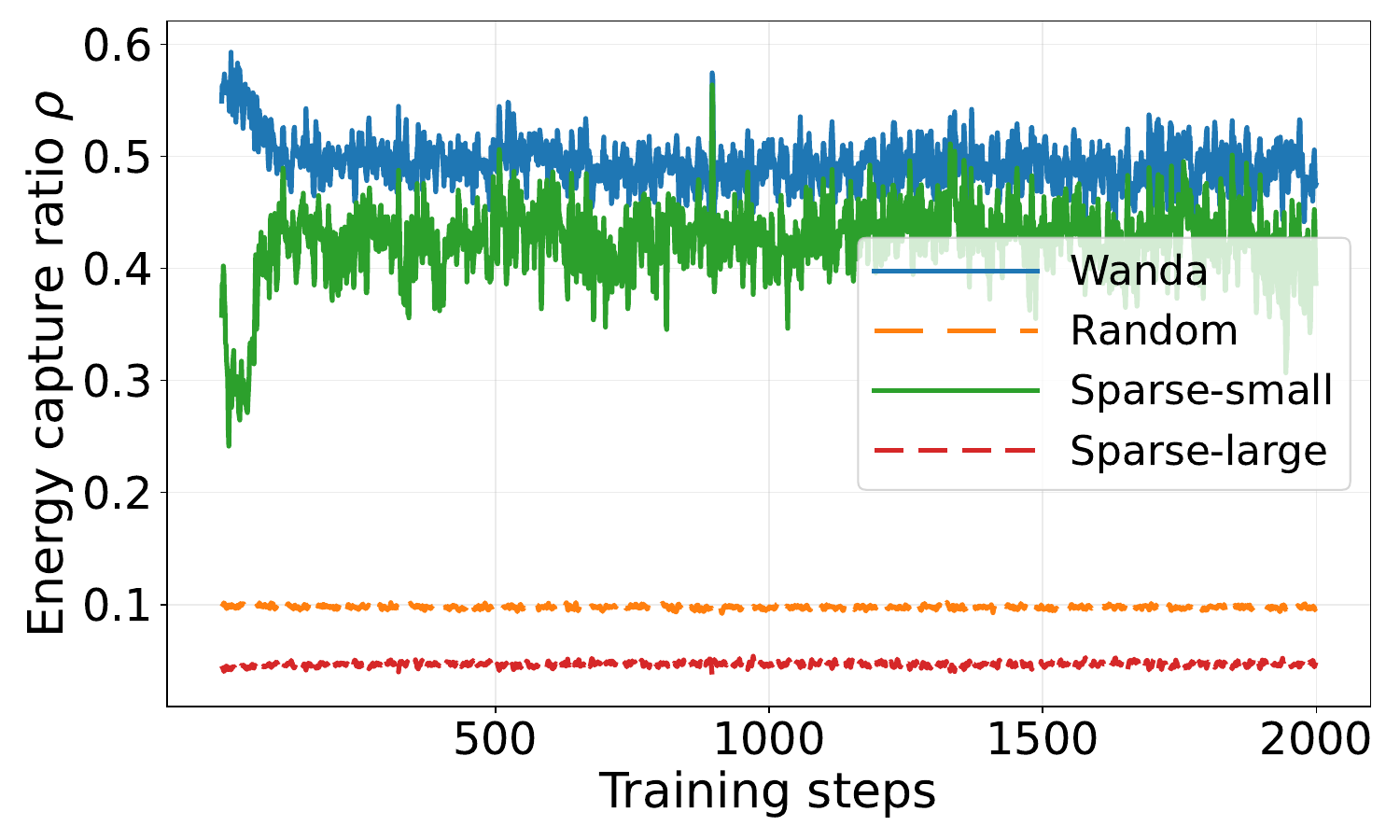
        }
        \caption{MNLI}
    \end{subfigure}

    \caption{
        \textbf{Gradient energy capture ratio $\rho$ on
        LLaMA-3.2-3B at a density ratio of $r=0.1$.}
        Random denotes uniform random masking. W-small and W-large
        select coordinates with the smallest and largest weight
        magnitudes, respectively, while WIS selects coordinates using
        Wanda weight--activation scores.
    }
    \label{fig:energy_llama_glue_010}
\end{figure*}

\begin{figure*}[!th]
    \centering

    \begin{subfigure}[b]{0.32\linewidth}
        \centering
        \includegraphics[width=\linewidth]{
            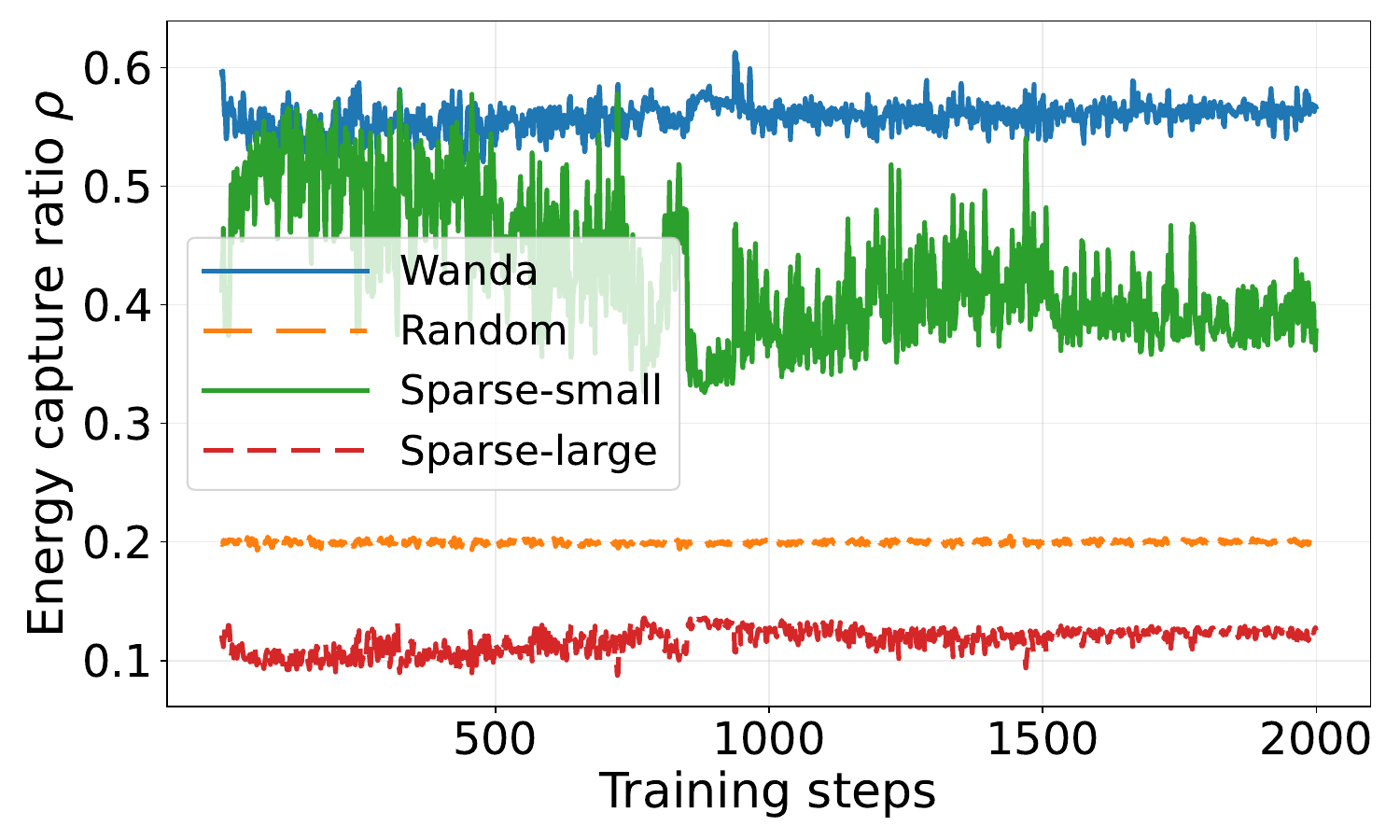
        }
        \caption{RTE}
    \end{subfigure}
    \hfill
    \begin{subfigure}[b]{0.32\linewidth}
        \centering
        \includegraphics[width=\linewidth]{
            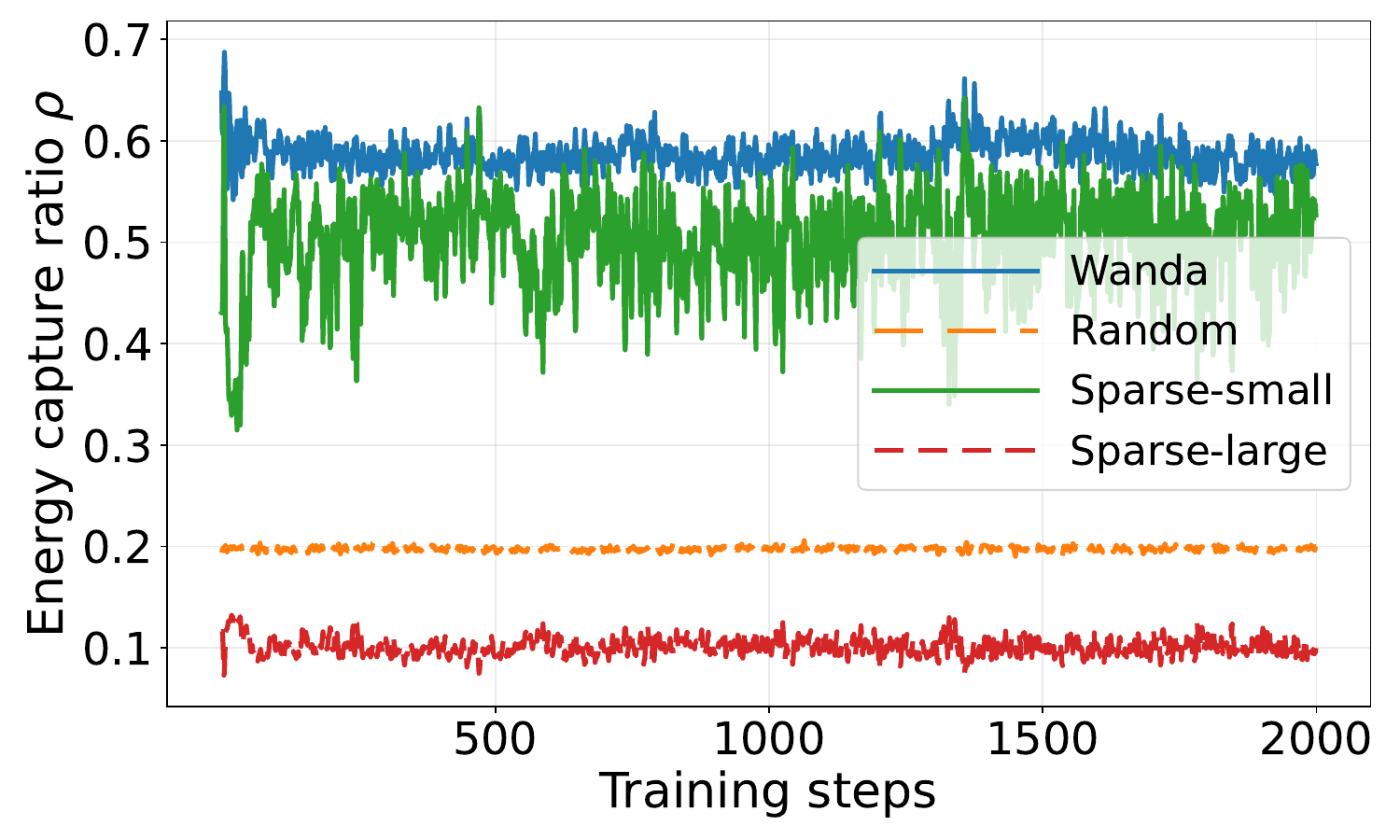
        }
        \caption{QQP}
    \end{subfigure}
    \hfill
    \begin{subfigure}[b]{0.32\linewidth}
        \centering
        \includegraphics[width=\linewidth]{
            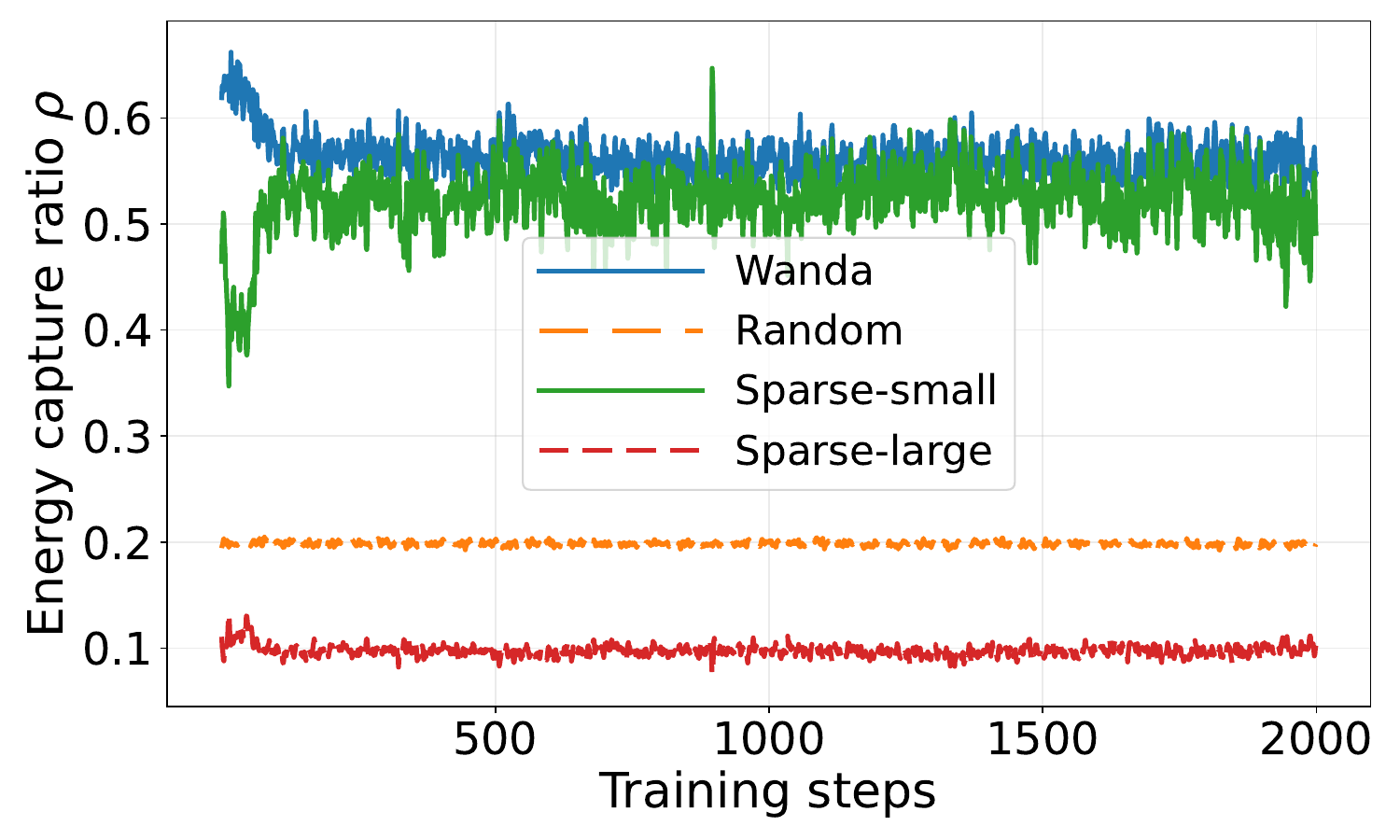
        }
        \caption{MNLI}
    \end{subfigure}

    \caption{
        \textbf{Gradient energy capture ratio $\rho$ on
        LLaMA-3.2-3B at a density ratio of $r=0.2$.}
        All mask construction strategies use the same row-wise
        perturbation density, model parameters, gradient snapshots,
        and evaluation data.
    }
    \label{fig:energy_llama_glue_020}
\end{figure*}

\begin{figure*}[!th]
    \centering

    \begin{subfigure}[b]{0.32\linewidth}
        \centering
        \includegraphics[width=\linewidth]{
            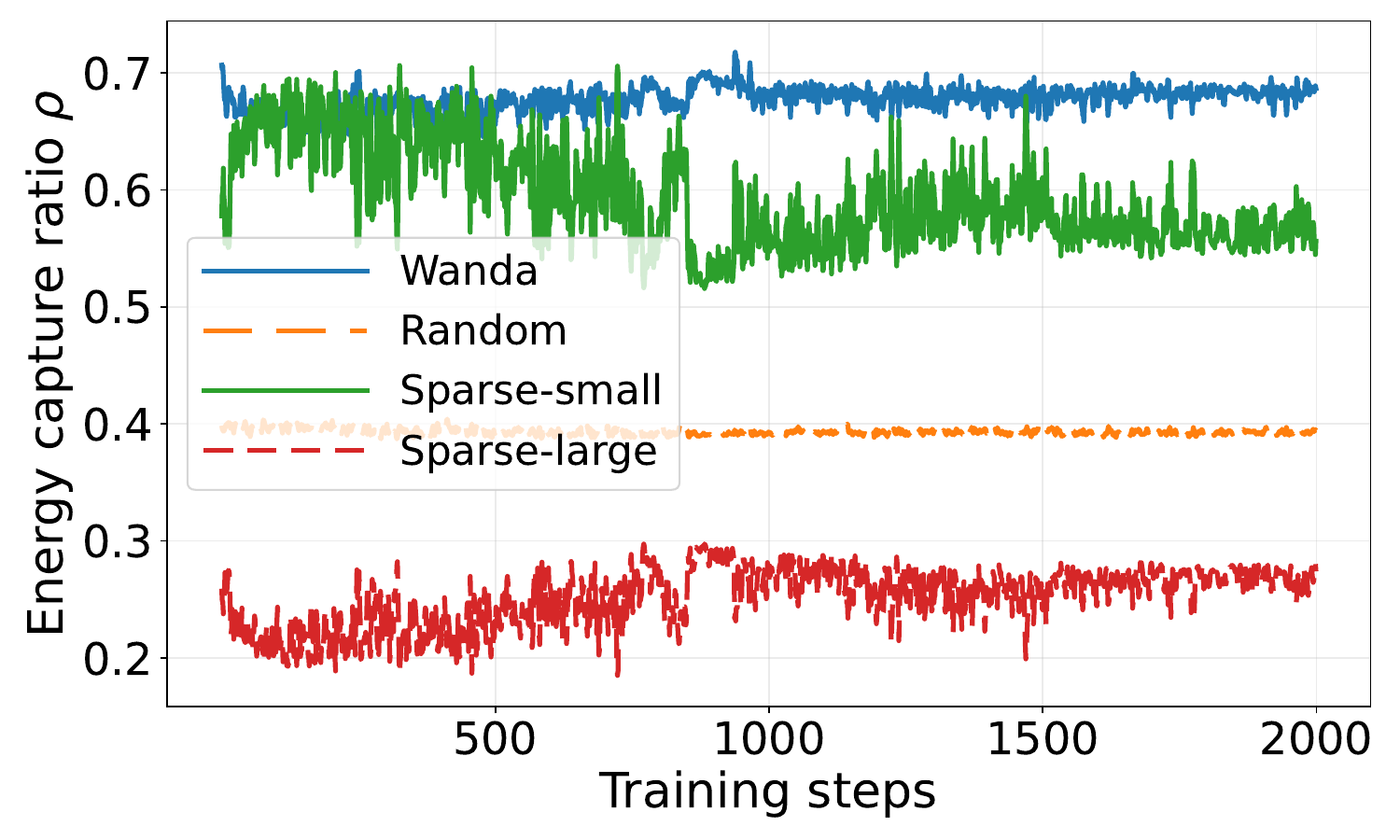
        }
        \caption{RTE}
    \end{subfigure}
    \hfill
    \begin{subfigure}[b]{0.32\linewidth}
        \centering
        \includegraphics[width=\linewidth]{
            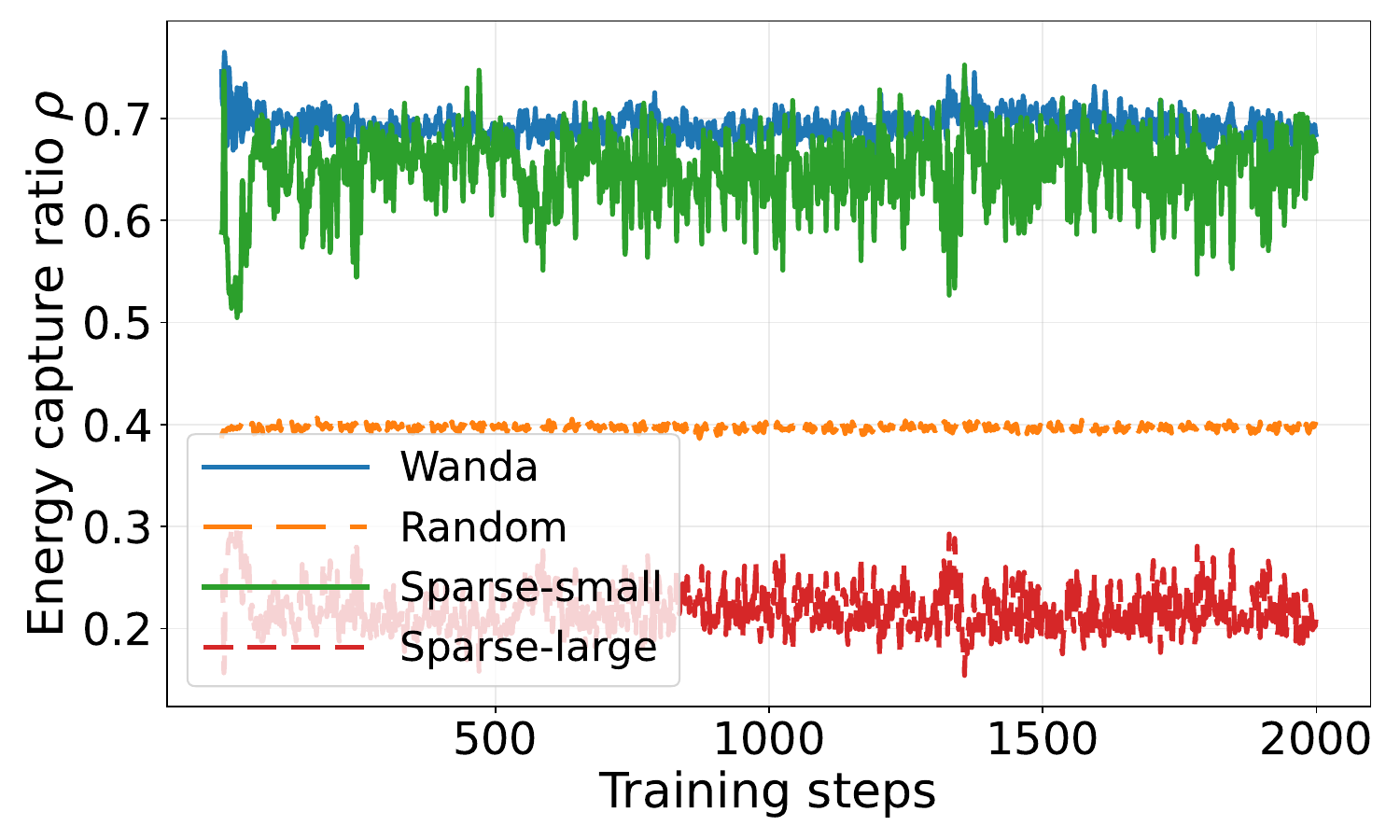
        }
        \caption{QQP}
    \end{subfigure}
    \hfill
    \begin{subfigure}[b]{0.32\linewidth}
        \centering
        \includegraphics[width=\linewidth]{
            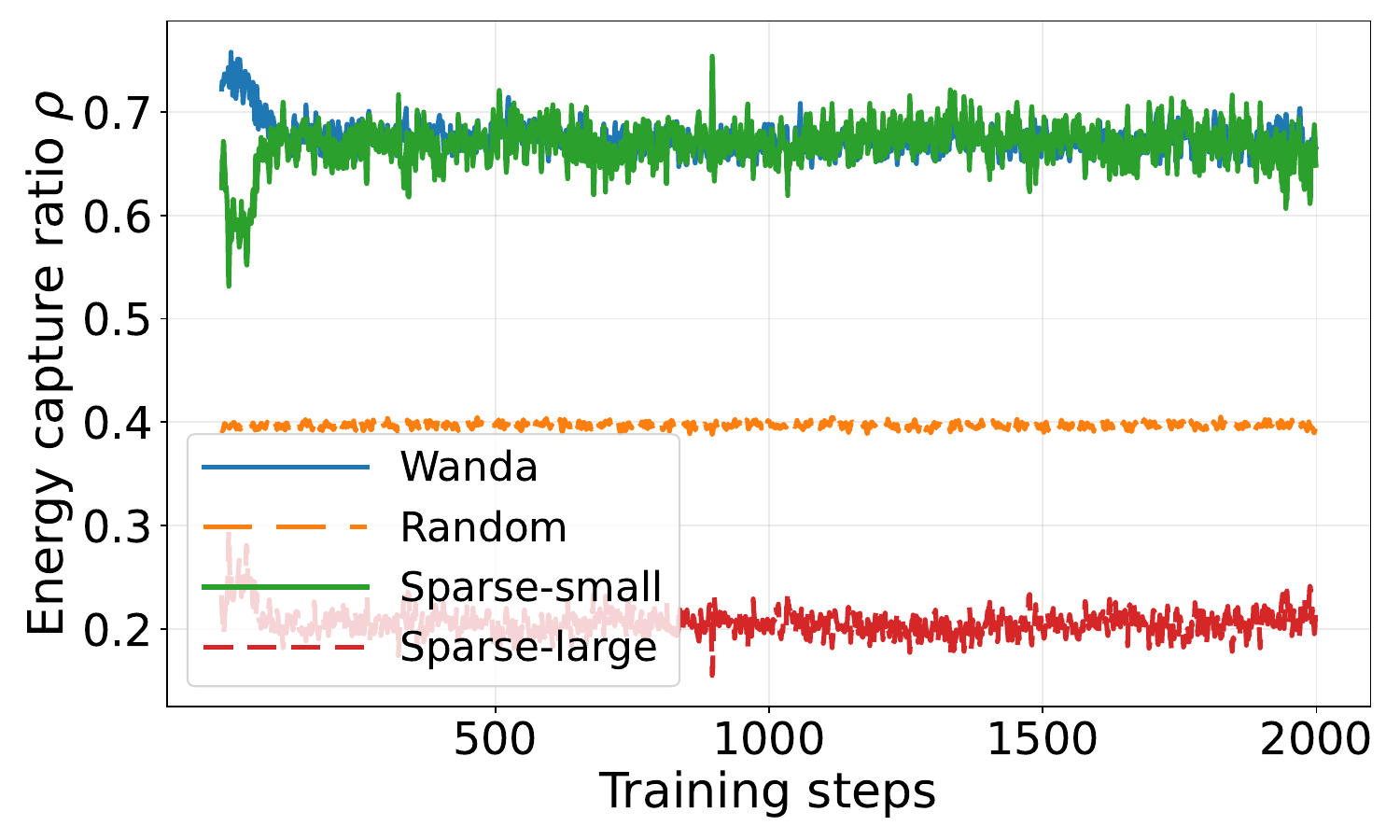
        }
        \caption{MNLI}
    \end{subfigure}

    \caption{
        \textbf{Gradient energy capture ratio $\rho$ on
        LLaMA-3.2-3B at a density ratio of $r=0.4$.}
        Random denotes uniform random masking. W-small and W-large
        select coordinates with the smallest and largest weight
        magnitudes, respectively, while WIS uses Wanda
        weight--activation scores.
    }
    \label{fig:energy_llama_glue_040}
\end{figure*}

Figures~\ref{fig:energy_llama_glue_010}--\ref{fig:energy_llama_glue_040}
report the additional results at density ratios of $0.1$, $0.2$, and
$0.4$, respectively.

WIS captures the most gradient energy in nearly all reported task--density settings. Its advantage is largest at low densities. At $r=0.4$ on MNLI, WIS and W-small have comparable capture ratios. W-small consistently captures more energy than W-large and generally exceeds random masking. These comparisons describe how the selected coordinates align with the shared AdamW gradient snapshots and show the contribution of downstream activation information to support selection.

\section{Detailed Pseudocode}
\label{app:pseudocode}

This section provides the detailed implementations of the parameter
perturbation and WIS mask construction routines used in
Algorithm~\ref{alg:wis-mezo}. For each linear layer,
$\mathbf{W}\in\mathbb{R}^{m\times n}$ denotes its weight matrix,
$\mathbf{Z}\in\mathbb{R}^{m\times n}$ denotes the random perturbation, and
$\mathbf{M}\in\{0,1\}^{m\times n}$ denotes the corresponding binary mask.

\begin{algorithm}[t]
\caption{Perturb Parameters}
\label{alg:perturbweight}
\begin{algorithmic}[1]
\STATE \textbf{Function}
$\operatorname{PerturbParameters}
(\mathbf{W},r,\epsilon,\mathbf{M},s_{\mathrm{rng}})$:
\STATE \textbf{Input:} Weights
$\mathbf{W}\in\mathbb{R}^{m\times n}$, density ratio $r$, perturbation scale
$\epsilon$, mask $\mathbf{M}\in\{0,1\}^{m\times n}$, and random seed $s_{\mathrm{rng}}$
\STATE Reset the random number generator with seed $s_{\mathrm{rng}}$
\STATE Sample
$\mathbf{Z}\in\mathbb{R}^{m\times n}$
with i.i.d.\ standard Gaussian entries
\STATE
$\mathbf{W}
\gets
\mathbf{W}
+
\frac{\epsilon}{\sqrt{r}}
\mathbf{Z}\odot\mathbf{M}$
\hfill \COMMENT{In-place perturbation}
\RETURN $\mathbf{W}$
\end{algorithmic}
\end{algorithm}

Seed replay regenerates the perturbation for the two loss evaluations and the update, retaining the memory advantages of MeZO.

\begin{algorithm}[t]
\caption{Compute Mask}
\label{alg:compute-mask}
\begin{algorithmic}[1]
\STATE \textbf{Function}
$\operatorname{ComputeMask}(\mathbf{W},\mathbf{X},r)$:
\STATE \textbf{Input:} Weights
$\mathbf{W}\in\mathbb{R}^{m\times n}$, calibration activations
$\mathbf{X}\in\mathbb{R}^{n\times B}$, and density ratio $r$
\STATE
$q_j\gets\|\mathbf{X}_{j,:}\|_2$,
for $j=1,\ldots,n$
\STATE
$\mathbf{S}_{i,j}\gets|\mathbf{W}_{i,j}|q_j$
\STATE
$k\gets\lfloor nr\rfloor$
\hfill \COMMENT{Selected parameters per row}
\FOR{each row $i$ in $\mathbf{S}$}
    \STATE
    $\mathcal{I}_i
    \gets
    \operatorname{TopK}(\mathbf{S}_{i,:},k)$
    \STATE
    $\mathbf{M}_{i,j}
    \gets
    \mathbb{I}[j\in\mathcal{I}_i]$,
    for $j=1,\ldots,n$
\ENDFOR
\RETURN $\mathbf{M}$
\end{algorithmic}
\end{algorithm}

Algorithm~\ref{alg:compute-mask} applies the Wanda score and selects the top-$k$ coordinates in each row. WIS-MeZO constructs the mask once, while WISD-MeZO refreshes it every $\tau$ steps.

\end{document}